\documentclass[11pt]{article}

\usepackage[margin=1in]{geometry}

\usepackage{microtype}
\usepackage{graphicx}
\graphicspath{{PaperICMLGraphics/}}
\usepackage{subcaption}
\usepackage{booktabs} 
\usepackage{multirow} 

\usepackage{authblk}

\usepackage{hyperref}
\hypersetup{colorlinks=true, linkcolor=blue, citecolor=blue, urlcolor=blue}

\usepackage{amsmath}
\usepackage{amssymb}
\usepackage{mathtools}
\usepackage{amsthm}

\usepackage[capitalize,noabbrev]{cleveref}

\usepackage[numbers,sort&compress]{natbib}

\theoremstyle{plain}

\theoremstyle{definition}

\theoremstyle{remark}

\usepackage[textsize=tiny]{todonotes}

\newcommand{\yrcite}[1]{\citeyear{#1}}

\begin{document}

\title{In-Context Molecular Property Prediction with LLMs: \\
A Blinding Study on Memorization and Knowledge Conflicts}

\author[1]{Matthias Busch\thanks{Corresponding authors: \texttt{matthias.busch@tuhh.de}, \texttt{christian.feiler@hereon.de}, \texttt{roland.aydin@hereon.de}}}
\author[3]{Marius Tacke}
\author[2]{Sviatlana V. Lamaka}
\author[2]{Mikhail L. Zheludkevich}
\author[1,3]{Christian J. Cyron}
\author[2]{Christian Feiler}
\author[1,4,5]{Roland C. Aydin}

\affil[1]{Institute for Continuum and Material Mechanics, Technical University of Hamburg, Eißendorfer Straße, 21073 Hamburg, Germany}
\affil[2]{Institute of Surface Science, Helmholtz-Zentrum Hereon, Max-Planck-Straße, 21502 Geesthacht, Germany}
\affil[3]{Institute of Material Systems Modeling, Helmholtz-Zentrum Hereon, Max-Planck-Straße, 21502 Geesthacht, Germany}
\affil[4]{German Center for Artificial Intelligence (DFKI), Saarland, Germany}
\affil[5]{Saarland University, Saarbrücken, Germany}


\date{}

\maketitle

\begin{abstract}
The capabilities of large language models (LLMs) have expanded beyond natural language processing to scientific prediction tasks, including molecular property prediction. However, their effectiveness in in-context learning remains ambiguous, particularly given the potential for training data contamination in widely used benchmarks. This paper investigates whether LLMs perform genuine in-context regression on molecular properties or instead rely on verbatim retrieval of memorized target values. Furthermore, we analyze the interplay between pre-trained knowledge and in-context information through a series of progressively blinded experiments. We evaluate nine LLM variants across three families (GPT-4.1, GPT-5, Gemini 2.5) on three MoleculeNet datasets (Delaney solubility, Lipophilicity, QM7 atomization energy) using a systematic blinding approach that iteratively reduces available information, complemented by 0-, 60-, and 1000-shot in-context sample sizes as an additional control for information access. To validate the memorization analysis and the blinding experiments, we add a positive and a negative control for the memorization experiments and structural reference baselines for the multi-shot experiments as well as bootstrap confidence intervals for all results. We find no evidence of verbatim retrieval on the legacy benchmarks and show that blinding exposes conflicts between pre-trained knowledge and in-context information. This work provides a principled framework for evaluating molecular property prediction under controlled information access.
\end{abstract}

\section{Introduction}

Large language models (LLMs) have emerged as powerful tools across numerous scientific domains, driven by their ability to process natural language and perform few-shot learning from minimal examples \cite{brown2020gpt3,dong2023survey,zhou2024mystery}. In chemistry, LLMs have been explored for diverse tasks \cite{NEURIPS2023_bbb33018} including molecule generation \cite{bran2023chemcrow} and property prediction \cite{jablonka2024leveraging,busch2025corrosion,joe2026icl}. The appeal of LLMs lies in their pre-trained knowledge and in-context learning (ICL) capabilities, which enable predictions with limited task-specific training data---a critical advantage in domains where experimental data is scarce or expensive to obtain.

However, the application of LLMs to molecular property prediction presents fundamental questions about the nature of their predictive capabilities. Unlike classification tasks, prediction requires precise numerical reasoning and an understanding of continuous relationships between molecular structure and properties. Recent work by Joe et al.~\yrcite{joe2026icl} has revealed that LLMs achieve near-perfect accuracy on raw molecular weight prediction from SMILES strings but fail when simple mathematical transformations are applied to the target values. This \textit{molecular weight anomaly}~\cite{joe2026icl} suggests that apparent success may stem from shortcut learning---exploiting superficial cues like token counting---rather than genuine structural reasoning.

\subsection{Motivation: Dataset Memorization and Benchmark Contamination}

A critical concern for evaluating LLMs on molecular property prediction is \textit{dataset contamination}: widely-used benchmark datasets like ESOL \cite{delaney2004esol}, Lipophilicity, and QM7 \cite{rupp2012qm7} have been publicly available for years and may appear in LLM training corpora. Recent studies have documented extensive memorization in LLMs \cite{carlini2023memorization,cheng2025survey} and raised concerns that high benchmark scores may reflect retrieval rather than learning \cite{sainz2023contamination,bordt2025forget}. When an LLM has encountered dataset labels during pre-training, strong in-context performance does not necessarily indicate genuine pattern recognition---it may simply trigger recall of memorized values.

This contamination problem is particularly acute for molecular property prediction, where the number of widely-used benchmarks is limited and repeatedly cited datasets (ESOL, MoleculeNet \cite{wu2018moleculenet}) dominate the literature. While such datasets provide valuable standardized evaluation, their ubiquity makes them poor tests of an LLM's ability to learn new functional relationships from in-context examples. This motivates our central approach: \textit{systematic information blinding} to test whether LLMs can generalize beyond memorization.

\subsection{Research Approach: Blinding and In-Context Benchmarking}

Inspired by the transformation framework of Joe et al.~\yrcite{joe2026icl}---which used mathematical transformations to prevent shortcut learning on molecular weight---we introduce a complementary \textit{six-level blinding approach} that progressively removes contextual information from the prediction task and transforms the target values. Our blinding levels range from full chemical context (Level 1: explicit property names, chemical terminology, untransformed values) to complete abstraction (Level 6: generic ``sample property'' prediction with transformed values and transformed SMILES strings, eliminating all domain context). This framework enables us to analyze the influence of different levels of prior knowledge on the molecular property prediction capabilities of LLMs.

As an initial investigation, we explore unblinded LLMs with varying amounts of in-context examples. We probe the models' prior knowledge via 0-shot prompting and test their in-context learning abilities with 60-shot and 1000-shot configurations. This experimental design serves dual purposes: (1) investigating potential conflicts between in-context sample information and prior knowledge, and (2) assessing the models' capability to extract structure-property relationships from large in-context datasets.

\subsection{Research Questions}

Our investigation addresses three key questions:

\begin{enumerate}
    \item \textbf{Memorization vs. domain knowledge}: To what extent do LLMs rely on dataset memorization versus genuine domain knowledge or general prediction capabilities? How does this balance shift across different molecular properties and varying levels of information blinding?

    \item \textbf{Interaction of prior knowledge}: How does an LLM's prior internal knowledge interact with provided in-context examples? Does strong prior knowledge facilitate learning, or does it hinder performance when it conflicts with the provided samples (e.g., in blinded settings)?

    \item \textbf{Implications for benchmarking and practice}: What are the broader implications for the use of standard molecular datasets as benchmarks, given the risks of contamination? Most importantly, what does this mean for researchers aiming to deploy LLM-based approaches for practical molecular property prediction?
\end{enumerate}

\section{Related Work}

\subsection{LLMs for Molecular Property Prediction}

Large language models have been applied to chemistry for various tasks \cite{NEURIPS2023_bbb33018} including molecule generation \cite{bran2023chemcrow} and property prediction \cite{jablonka2024leveraging}. While chemistry-specific models like ChemLLM \cite{zhang2024chemllm} have been developed, our recent work Busch et al.~\yrcite{busch2025corrosion} shows that general-purpose LLMs (GPT-4o, o1) with strong reasoning outperform domain-specialized models on regression tasks, motivating our focus on general LLM families.

For in-context learning (ICL) on regression, Akyürek et al.~\yrcite{akyurek2022transformers} demonstrate transformers can implicitly implement ridge regression. Critically, Joe et al.~\yrcite{joe2026icl} reveal LLMs exploit shortcuts in molecular weight prediction, achieving near-perfect accuracy on raw values but failing under nonlinear transformations---motivating our combined blinding and transformation approach. Building on the former result, a fully blinded LLM can be viewed as performing automated featurization followed by in-context regression---a learned analogue of the classical recipe of mechanical featurization (e.g., bag-of-characters counts) followed by ridge regression. 

\subsection{Memorization, Contamination, and Context Effects}

Carlini et al.~\yrcite{carlini2023memorization} demonstrate LLMs verbatim reproduce training data, with memorization increasing with model size. Widely used benchmarks like ESOL \cite{delaney2004esol} from MoleculeNet \cite{wu2018moleculenet} are vulnerable to contamination \cite{chen2025benchmarking}. Our blinding approach addresses this by removing contextual information and applying value transformations, complementing the mathematical transformation framework of Joe et al.~\yrcite{joe2026icl}.

Liu et al.~\yrcite{liu2024lost} show LLMs struggle with the ``lost in the middle'' phenomenon in long contexts. Our evaluation with up to 1000 examples tests whether LLMs extract functional relationships from large datasets.

\subsection{Positioning}
\label{sec:positioning}

To our knowledge, this is the first work that systematically evaluates the in-context-learning (ICL) performance of LLMs for molecular property regression under controlled information removal (blinding). We further believe this is the first systematic study to probe LLM ICL performance on large \textit{experimental} datasets with up to 1000 ICL samples, providing enough data to enable predictions even without prior knowledge. Earlier studies, such as Joe et al.~\yrcite{joe2026icl}, use orders of magnitude fewer samples for ICL and computed rather than experimental quantities. Jablonka et al.~\yrcite{jablonka2024leveraging} and our own previous study, Busch et al.~\yrcite{busch2025corrosion}, also use fewer ICL samples. None of these works systematically removed information from the context window to analyze its influence. In this work, we analyze which capabilities LLMs rely on when learning in-context and predicting molecular properties. For this, we use models from three LLM families across three popular benchmark datasets. We chose the datasets such that we expect the LLMs to know them because they all predate 2018. We do this because we want to also see memorization effects if they do exist. To put these effects in context, we additionally use control datasets: one created after the models' knowledge cutoff date, and two small datasets drawn from sources guaranteed to be in the training corpus.

\section{Methodology}

\subsection{Datasets}

We evaluate on three MoleculeNet benchmark datasets~\cite{wu2018moleculenet}: ESOL~\cite{delaney2004esol} measures aqueous solubility for 1,128 organic compounds (log mol/L), Lipophilicity contains octanol-water partition coefficients (logD at pH 7.4) for 4,200 compounds, and QM7~\cite{rupp2012qm7,huggingfaceqm7} provides DFT-computed atomization energies (kcal/mol) for 6,834 small molecules. We emphasize that for ESOL we use the \textit{experimentally measured} log-solubility column as the target; the dataset's \textit{calculated} ESOL value (the output of Delaney's regression equation) is used neither as a target nor as an input feature. Likewise, the Lipophilicity target is the \textit{experimental} logD at pH~7.4; the model only ever receives the SMILES string, never a computed descriptor. Note that we use a HuggingFace variant of QM7 with SMILES strings as sole descriptors rather than the standard version with 3D coordinates, resulting in slightly fewer samples; as a consequence its structure--property statistics differ from the canonical 3D QM7 (see the Appendix). These datasets provide complementary challenges: experimentally measured physicochemical properties (ESOL, Lipophilicity) versus quantum-mechanical predictions (QM7), with varying dataset sizes and chemical diversity. In all cases, the only input provided to the model is the SMILES string~\cite{weininger1988smiles} of the molecule, together with the target property value.

Additionally, we used a set of 41 IUPAC atomic masses and 45 boiling points reported on Wikipedia as a positive memorization control, and the 2025 ASAP~Discovery antiviral potency dataset from the ASAP\,$\times$\,Polaris\,$\times$\,OpenADMET blind challenge~\cite{asap_polaris_openadmet_2025} as a negative control.

\subsection{Six-Level Blinding Framework}

\begin{table*}[t]
\centering
\small
\caption{Six-level blinding framework. Each level progressively reduces contextual information to isolate different LLM capabilities. ``Label'' indicates if target values are mathematically transformed; ``Input'' indicates if SMILES strings are transformed.}
\begin{tabular}{clccp{5.5cm}}
\toprule
Level & Name & Label & Input & Context description \\
\midrule
1 & Specific & Original & Original & \multirow{2}{5.5cm}{Full chemical context: property name (e.g., ``log solubility''). Tests memorization + domain knowledge.} \\
2 & Specific-Transformed & Transformed & Original & \\ \addlinespace[8mm]
3 & Generic & Original & Original & \multirow{2}{5.5cm}{Generic ``molecular property'' label. Tests chemical in-context learning without property-specific priors.} \\
4 & Generic-Transformed & Transformed & Original & \\ \addlinespace[8mm]
5 & Agnostic & Original & Transformed & \multirow{2}{5.5cm}{Domain-agnostic framing (``sample property''). Tests general in-context learning.} \\
6 & Agnostic-Transformed & Transformed & Transformed & \\ \addlinespace[4mm]
\bottomrule
\end{tabular}
\label{tab:blinding_levels}
\end{table*}

To test whether LLMs perform genuine in-context learning versus memorization or shortcut exploitation, we introduce a systematic \textit{information blinding} approach that progressively reduces contextual information. Our framework comprises six levels (Table~\ref{tab:blinding_levels}).

The ``specific'' levels specify the molecular property (e.g., ``exact'' = ``log solubility'' or ``close'' = ``property related to solubility''), allowing LLMs to apply domain knowledge and use memorized values. Levels 3-4 use generic ``molecular property'' labels, preventing domain-specific reasoning while retaining chemistry terminology. Levels 5-6 remove all chemistry context (``samples'', ``structure strings'' instead of ``molecules'', ``SMILES'') and transform SMILES strings (see Section~\ref{sec:transformation}), forcing pattern recognition on an unfamiliar structural language. Levels 2, 4, and 6 apply the transformation described in Section~\ref{sec:transformation} to reduce the impact of memorized values. Detailed prompts and blinding descriptions are in Appendix.

To interpret our findings and their implications, we introduce a framework distinguishing four levels of LLM capabilities in molecular property prediction. Each level corresponds to a specific level of generalization. With blinding the dataset and the prompt, one can deactivate certain capabilities, and by comparing the results of different blinding levels, one can qualitatively analyze the influence of the different capabilities on the overall performance.

\begin{enumerate}
    \item \textbf{Direct memorization}: The ability to exactly reproduce a value from training data without generalization. (Used by: blinding level 1, and possibly 3 and 5).
    \item \textbf{Learned structure-property relationships}: The ability to approximate values for known properties using learned correlations from specified input data like SMILES strings. (Used by: blinding levels 1-2).
    \item \textbf{Chemical In-Context Learning}: The ability to predict unknown properties for a given molecular structure by applying general chemical principles and extracting structure-property relations from molecular in-context examples. (Used by: blinding levels 1-4).
    \item \textbf{General In-Context Learning}: The ability to predict unknown properties solely from unknown structural patterns (e.g. SMILES syntax, but transformed) and in-context examples, without relying on chemical priors or property labels. (Used by: blinding levels 1-6).
\end{enumerate}

This taxonomy allows us to dissect the ``performance'' of LLMs not as a monolithic metric, but as a composite of these distinct mechanisms. It is important to note that these capabilities are not mutually exclusive, nor do they combine linearly, and experiments have to be designed carefully to disentangle them. In addition to the blinding levels, using 0-shot deactivates capabilities 3 and 4 since they rely on generalization from in-context examples while capabilities 1 and 2 rely only on the model's pre-trained knowledge and the provided input data.

\subsection{Transformation Functions}
\label{sec:transformation}

For blinding levels 2, 4, and 6, we apply a single mathematical transformation to the target values, which inverts the labels (multiplication by $-1$) and rescales them to the range $[0, 100]$. The inversion disrupts value memorization, while the rescaling prevents simple recognition of the original scale; both still preserve the qualitative relationship between input and label. This single transformation is applied consistently across all three blinded levels and all datasets. We report all metrics on inverse-transformed predictions to enable fair comparison across blinding levels. The benefit of a linear transformation is that it keeps the base difficulty of prediction exactly the same while preventing direct verbatim recognition of the values in the dataset. As a control experiment, we also ran a non-monotonic double-sine transformation, which led to similar results (see the Appendix). 

For blinding levels 5 and 6, we additionally transform the SMILES strings themselves to prevent the LLM from recognizing familiar chemical structures. We apply a deterministic character substitution where each atom symbol or SMILES syntax character is replaced by a unique substitute character. For example, atoms like C, N, O, Cl are mapped to arbitrary characters, and SMILES syntax elements like parentheses, brackets, and bond symbols are similarly replaced. This transformation is applied consistently across all molecules in both training and test sets. Critically, this substitution preserves the structural relationships encoded in the SMILES syntax---branching, ring closures, and connectivity remain intact---while rendering the strings unrecognizable to the LLM. This approach is analogous to the linear transformation of labels: it maintains the structure-property relationships that can be learned from in-context examples while disabling direct lookup of memorized chemical knowledge.

\subsection{Prompting Strategy}

We employ the pre-analysis prompting approach of Busch et al.~\yrcite{busch2025corrosion}, which separates prediction into two phases: (1) Analysis: systematically examine training data to identify structural features, patterns, and similar molecule pairs. (2) Prediction: find similar training molecules to the test case, rank by similarity, and compute a weighted average prediction. This structured approach encourages chain-of-thought reasoning~\cite{wei2022chain}. Furthermore, for the 0-shot experiments, we apply a simple prompt that asks the model to predict the target value without reasoning. Details of the prompts are provided in Appendix.

\subsection{In-Context Learning Setup}

We evaluate three configurations: 0-shot (no training examples), 60-shot (60 random training molecules), and 1000-shot. We use the same random train-test split with 150 test samples for all runs except the 0-shot runs (1000 test samples) to balance statistical robustness. We perform 2 runs per configuration and dataset to limit the computational cost.

\subsection{Evaluation Metrics}

We evaluate prediction quality using the Pearson correlation coefficient ($r$), since MAE and RMSE would be affected by the transformation's scale shift and by the differing scales of the dataset labels, whereas the correlation is invariant under such rescaling. Only the cumulative error distributions use MAE as the single-sample error metric. All metrics are computed on a held-out test set not included in in-context training examples.

\subsection{Implementation Details and Large Language Models}

We evaluate nine LLM variants across three families and three size classes: GPT-4.1\linebreak (\texttt{gpt-4.1-nano}, \texttt{gpt-4.1-mini}, \texttt{gpt-4.1}) and GPT-5 (\texttt{gpt-5-nano}, \texttt{gpt-5-mini}, \texttt{gpt-5}) by OpenAI, and Gemini 2.5 (\texttt{gemini-2.5-flash-lite}, \texttt{gemini-2.5-flash}, \texttt{gemini-2.5-pro}) by Google. This $3 \times 3$ design enables comparison of both family-specific capabilities and size-dependent scaling effects. The LLMs are accessed via Azure (GPT models) and OpenRouter (Gemini models) with sampling parameters (temperature = 0.7, top\_p = 0.95), where available. GPT-5 family models do not allow setting the temperature and top\_p parameters.

\section{Results}

\begin{figure*}[!htbp]
\begin{center}
\includegraphics[width=\textwidth]{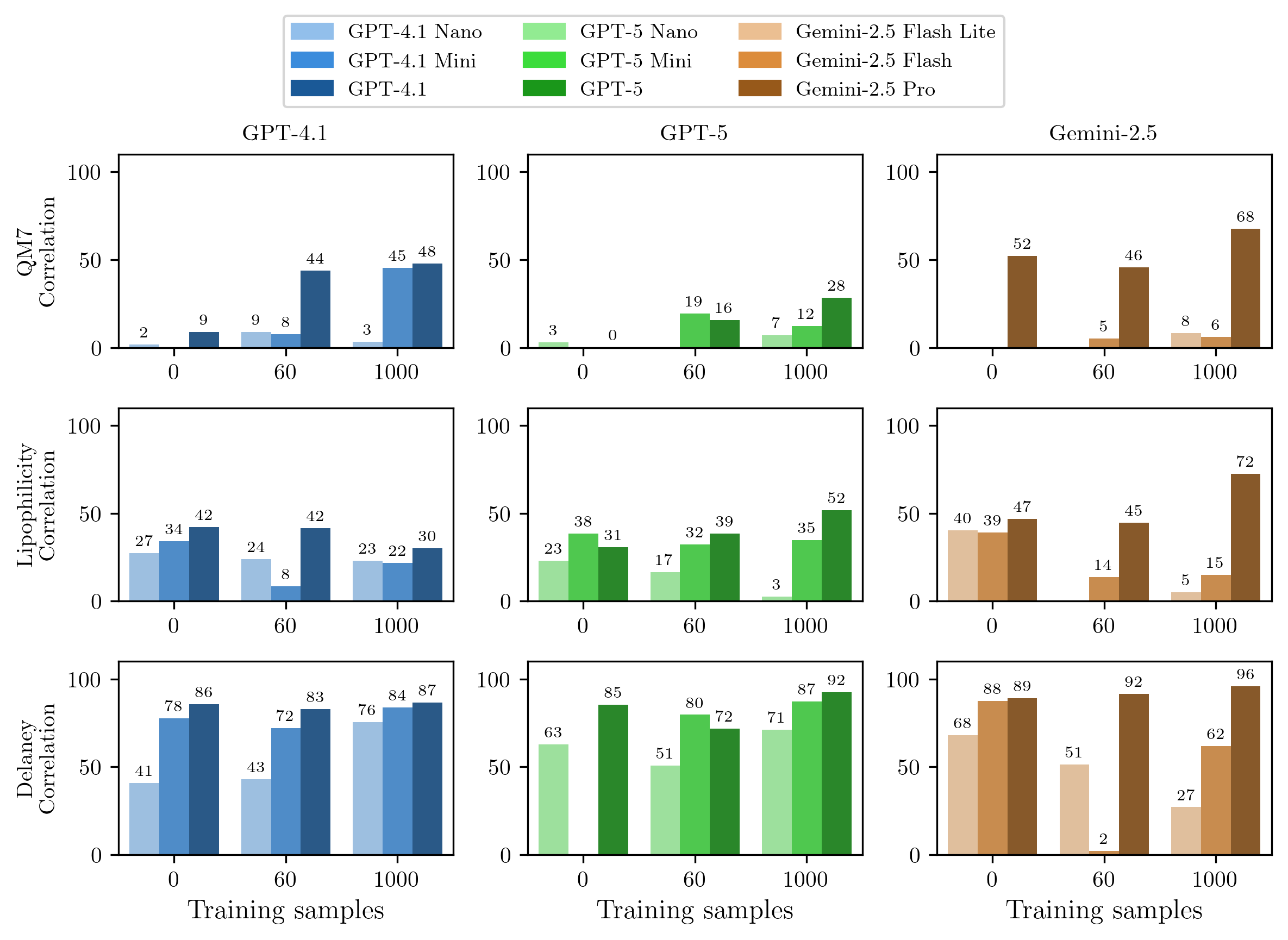}
\caption{Correlation coefficients for molecular property prediction across three datasets (Rows: QM7, Lipophilicity, Delaney) and three LLM families (Columns: GPT-4.1, GPT-5, Gemini 2.5). The x-axis represents the input configuration (0-shot, 60-shot, 1000-shot). Different colors indicate model sizes within each family. Higher correlation values indicate better performance. Note the varying 0-shot correlations across the different datasets, indicating the popularity of the datasets in the training corpus, while they are relatively constant across the different model families. Also note that the 1000-shot correlation is usually higher than the 0-shot correlation, while the 60-shot correlation varies.}
\label{fig:correlation_results}
\end{center}
\end{figure*}

\subsection{General Performance and In-Context Learning Scaling}

Figure~\ref{fig:correlation_results} displays correlation coefficients across three LLM families, three datasets, and three input configurations (0, 60, and 1000 shots) without blinding. While the largest models generally perform well given that they only receive SMILES strings, they do not surpass published baselines like Chemprop \cite{yang2019analyzing} or MPNNs \cite{gilmer2017message}.

However, the purpose of these experiments is to check the pre-trained capabilities (memorization and learned relationships) with the 0-shot experiments and possible improvements with in-context learning. The 0-shot performance is mixed for the different datasets but relatively consistent over the different models (with some exceptions). For QM7, the 0-shot performance is around 0 for all models except Gemini 2.5 Pro with around $53\%$, for Lipophilicity, the 0-shot performance is around $30-40\%$ for all models and for Delaney, the 0-shot performance is around $85-90\%$ for the largest models and decreases for smaller models. Since the 0-shot performance measures the pre-trained capabilities and completely ignores in-context learning, the results are a proxy for the prevalence of the respective dataset or the general underlying property in the models' training corpus. Hence, our results indicate that the Delaney dataset or solubility is the most popular among the three, followed by Lipophilicity and then QM7 being seemingly absent from the models' effective training data. It should be noted that the variant of QM7 we use here provides SMILES strings rather than the standard 3D representation, which might also be a reason why most models do not recognize it.

When adding samples and in-context learning, the performance changes differently for the different models. The largest models behave similarly for all datasets: their performance increases in 8/9 cases from 0 to 1000 shots. However, the performance in 60 shots is in about half of the cases worse than the 0-shot performance. This indicates that the models struggle to combine their internal knowledge with a small number of samples but improve once a critical mass of samples is provided. Although individual per-model differences fall within the test-set confidence interval, these directional trends are jointly significant across the nine models by a sign test (1000-shot above 0-shot for the flagship models, $p=0.02$; 60-shot below 0-shot, $p=0.026$; 1000-shot above 60-shot, $p=0.003$; Appendix, significance analysis). Middle and small-sized models show inconsistent performance. Gemini Flash and Flash-Lite clearly decrease or stagnate near 0 with more samples. OpenAI's mini and nano models are more robust; while their 0-shot average is lower, they handle additional samples effectively, maintaining performance without the degradation seen in the Gemini models.

\begin{figure}[!htbp]
\begin{center}
\includegraphics[width=0.5\textwidth]{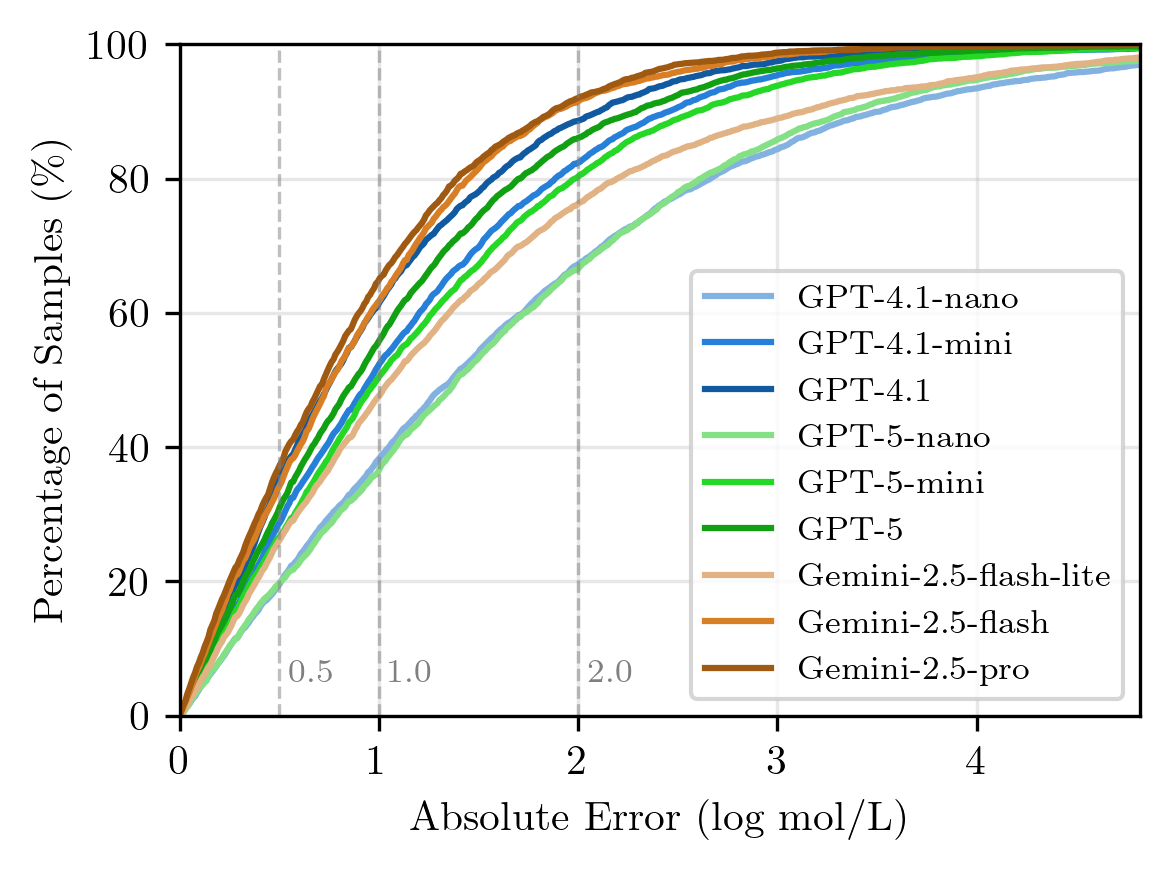}
\caption{Cumulative error distribution for 0-shot predictions on the Delaney dataset. The x-axis shows the absolute error threshold, and the y-axis shows the percentage of samples with error below that threshold. Steeper curves indicate better performance. Note the continuous distribution of errors without a substantial number of errors at zero, which would be the expected behavior if the LLMs had memorized the target values.}
\label{fig:cumulative_error}
\end{center}
\end{figure}

\subsection{Error Distribution and Memorization Analysis}
\label{sec:memorization_results}

Figure~\ref{fig:cumulative_error} illustrates the cumulative error distribution for 0-shot predictions on the Delaney dataset. As can be seen, there is a continuous distribution of errors without a substantial amount of errors at zero. If the models had memorized the target values, we would expect a significant fraction of samples with exactly zero error. Since this is not the case, we conclude that the models do not reproduce the target values verbatim. This is consistent with the models relying on a learned relationship between SMILES strings (or molecular structures) and solubility, though it does not by itself exclude approximate recall or other statistical associations acquired during pre-training (see the Discussion).

To quantify potential memorization effects in greater detail, we analyzed the exact memorization of sample properties by the LLMs for the three legacy datasets we use throughout our study: Delaney, Lipophilicity, and QM7. Additionally, we included three control datasets: On the one hand a post knowledge cutoff dataset as negative control for retrieval---the 2025 ASAP~Discovery antiviral potency dataset---and on the other hand two additional datasets as positive control where direct verbatim retrieval is very likely---a collection of 41 IUPAC standard atomic weights and a set of 45 boiling points from Wikipedia. 

For all 6 datasets, we report the exact match rate (first three significant digits), two precision-scaling \textit{retention} ratios ($m_2/m_1$ and $m_3/m_2$) that measure the fraction of matches at one precision that survive to the next significant digit, and the median Pearson correlation of the per-molecule mean prediction. Because a value reported to one or two digits can match trivially, all match rates are computed over ground-truth values carrying at least three significant digits. A bare match rate, however, only catches \textit{lossless} recall; Table~\ref{tab:memorization} therefore pairs it with a precision-scaling retention statistic.

\begin{table}[h]
    \centering
    \caption{Memorization diagnostics (0-shot, pooled over the nine models). For each dataset we report the three-significant-digit exact-match rate; the two precision-scaling \textit{retention} ratios $m_2/m_1$ and $m_3/m_2$---the fraction of matches at one precision that survive to the next significant digit, with coincidence floor $\approx10\%$ and genuine recall $\approx100\%$ (the lower-precision $m_2/m_1$ rests on roughly $10\times$ more matches and is the statistically tighter estimate); and the median Pearson $r$ of the per-molecule mean prediction. Note that the three legacy benchmarks' results match qualitatively with the 2025 post-cutoff antiviral control, all sitting at the coincidence floor on \textit{both} ratios together with near-zero match rates, whereas the positive control of demonstrably memorized values (IUPAC standard atomic weights; English-Wikipedia boiling points) is near-total on both under the \textit{identical} detector. Detailed per-model counts, the full per-model retention tables, and the positive control are in the Appendix.}
    \label{tab:memorization}
    \small
    \begin{tabular}{l|rrrr}
        \toprule
        \textbf{Dataset} & \textbf{3-sig match \%} & \textbf{$m_2/m_1$} & \textbf{$m_3/m_2$} & \textbf{Median $r$} \\
        \midrule
        Delaney (ESOL)                  & 0.33     & 10.8\% & 12.2\% & 0.78 \\
        Lipophilicity                   & 0.33     & 11.6\% & 11.0\% & 0.39 \\
        QM7                             & 0.16     & 17.8\% & 7.6\%  & 0.01 \\
        Antiviral potency (post-cutoff) & 0.13     & 9.3\%  & 11.3\% & 0.37 \\
        \midrule
        \textit{Atomic weights} (memorized) & 95--100 & 99.9\% & 97.7\% & 1.00 \\
        \textit{Boiling points} (memorized) & 16--78  & 81.7\% & 83.8\% & 0.88 \\
        \bottomrule
    \end{tabular}
\end{table}

Table~\ref{tab:memorization} shows the results of our 0-shot experiments, pooled across all models (detailed results in the Appendix). One can clearly see three times the same pattern. First, looking at the bare 3-significant-digit match rates, the results are split between the positive control datasets and the negative control. The models can recall IUPAC atomic weights almost with $100\%$ certainty and Wikipedia boiling points mostly above $50\%$ (see the Appendix). On the other side sit the Antiviral potency dataset as negative control with near zero match rate ($0.13\%$). The three legacy benchmarks are similar in match rate to the negative control, with near-zero match rates ($0.16$--$0.33\%$). Second, the precision-scaling retention ratios show a similar split: The positive controls are close to $100\%$ retention, while the three legacy benchmarks and the post-cutoff antiviral control are all around the $\sim$10\% coincidence floor at \textit{both} precision steps. Third, the median Pearson correlation shows this split as well, but on a much more continuous scale, with the three legacy benchmarks and the negative control showing a range of $0.01$--$0.78$ and the positive controls showing near-perfect correlation ($0.88$--$1.00$). 

First of all, the positive control shows how LLMs perform on a dataset where they have demonstrably memorized the values. The results show large fractions of 3-significant digit matches, with near-total retention at both precision steps. On the other side, the negative control of the 2025 post-cutoff antiviral potency dataset shows that even on an unseen dataset, the models can achieve a non-trivial correlation ($0.37$) and a minimal exact-match rate ($0.13\%$). However, the retention ratios show clearly that the correlation is based on the first significant digit, with the second and third digits being essentially randomly distributed. With one of ten possible digits matching by chance, i.e.\ $\approx10\%$ (for QM7 the second digit is not perfectly evenly distributed; see the Appendix). The three legacy benchmarks show the same pattern as the negative control, which is why we conclude that the models probably do not retrieve the values of these specific datasets verbatim. While correct prediction of a single digit in $40$-$50\%$ of the cases could theoretically be attributed to lossy recall of a memorized value, this rather looks like a very coarse approximation of the underlying chemistry, with the correct order of magnitude and leading digit being inferred from the structure. 

We stress that this analysis does not prove that the datasets are absent from training (that would require access to the undisclosed corpora), but it provides strong evidence that the models have at least not memorized large fractions of their values during training. It also does not rule out approximate (lossy) recall, substructure-level associations, or other statistical regularities absorbed during pre-training, which may still contribute to the 0-shot predictions. But although the three legacy datasets clearly predate the development of the LLMs, they show the same characteristics as the dataset that was certainly not part of the training corpus. The 0-shot correlation instead reflects how strongly the underlying property was represented in the training corpus.

As a test of whether our blinding protocols can suppress recall, we applied them to the positive control datasets, albeit with a different prompting strategy (see SI for full details). The results show that the stronger LLMs can uncover an inverted and rescaled property from in-context examples when they have memorized the underlying property and are given the property name. They can also find the underlying property if the name is masked. However, combining both the masking of the property name and the transformation of the target values suppresses a large part of the recall, but not completely. The Agnostic blinding levels, which additionally transform the structure representation, bring recall down to the coincidence floor: retention there is statistically indistinguishable from chance, although the point estimates leave room for a small residual leakage that 45 compounds cannot resolve. The blinding is therefore effective at suppressing recall even on a dataset the models have clearly memorized, but only in its fully blinded form: the property name alone protects nothing, and the value transformation protects only against the weaker models. 

\subsection{Effect of Blinding and Information Removal}

\begin{figure}[!htbp]
\begin{center}
\includegraphics[width=0.5\textwidth]{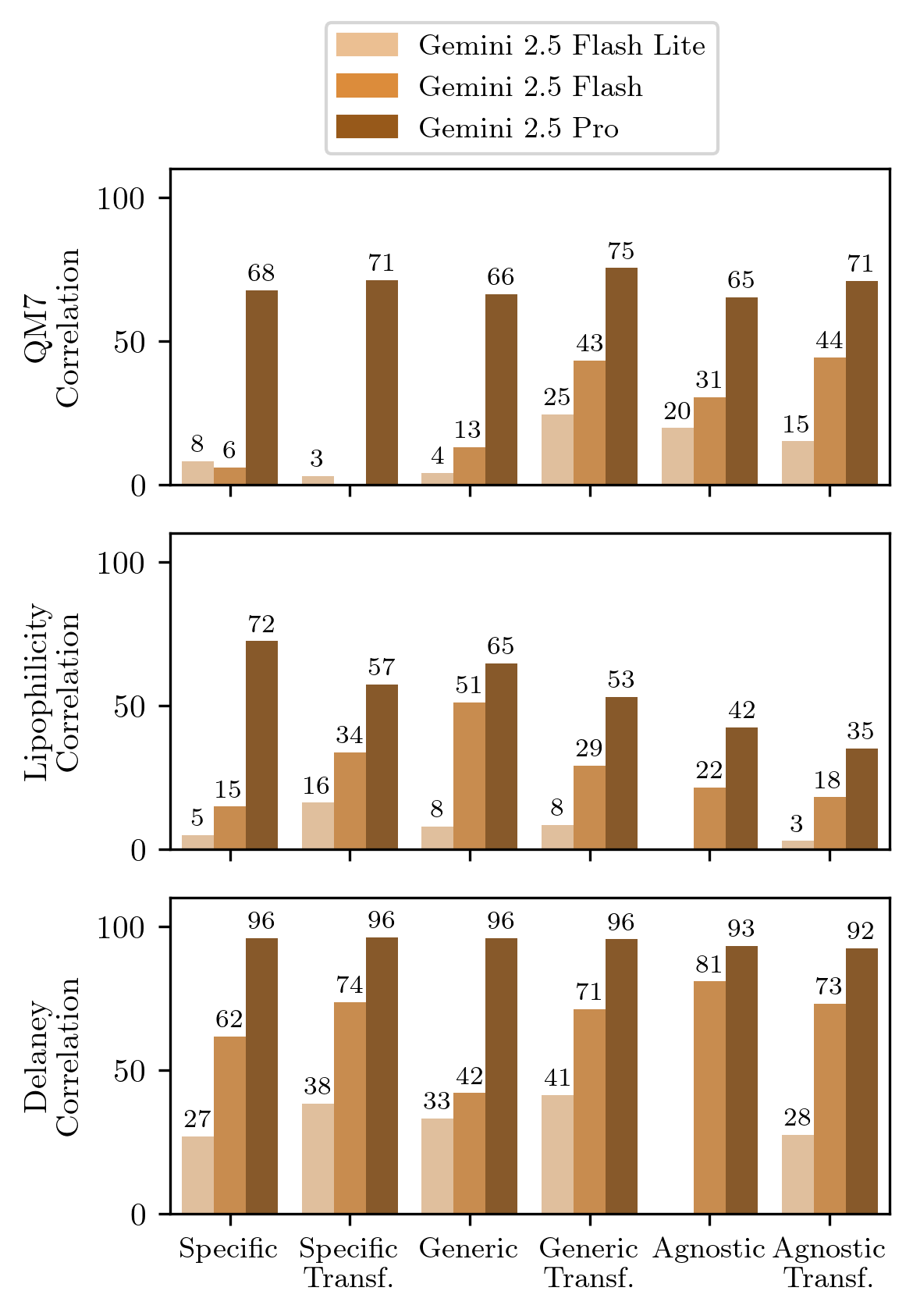}
\caption{Detailed performance breakdown for Gemini 2.5 models across all six blinding levels (x-axis) and three datasets (rows). Specific blinding levels specify the property to be predicted by name, while generic blinding levels name it as molecular property and agnostic blinding levels name it as sample property. ``Transf.'' relates to the label values being transformed as described in the Methods section. Note that the peak correlation value is frequently not at the first blinding level or approximately shared across different blinding levels.}
\label{fig:gemini_approaches}
\end{center}
\end{figure}

\begin{figure*}[!htbp]
\begin{center}
\includegraphics[width=\textwidth]{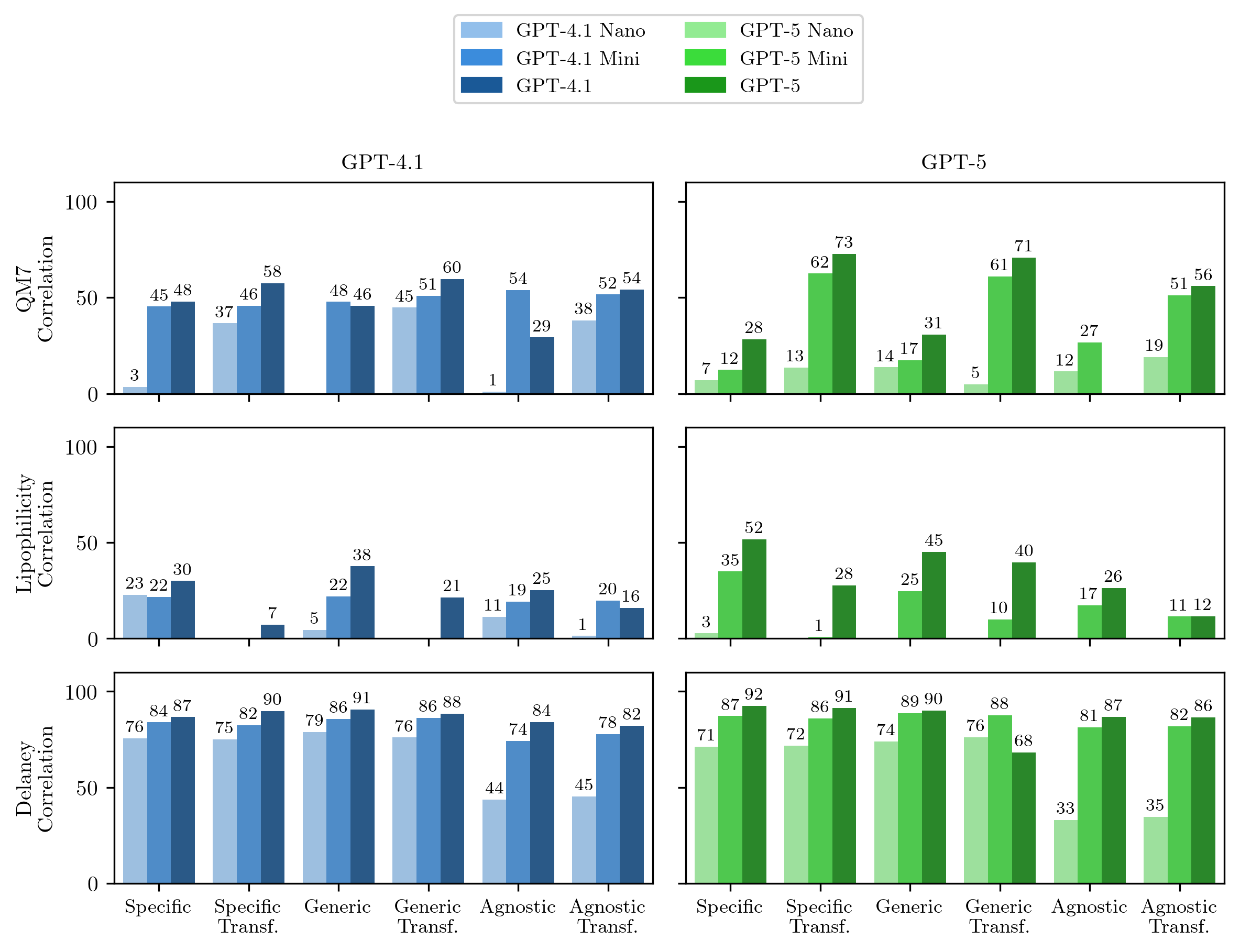}
\caption{Correlation of OpenAI models (GPT-4.1 and GPT-5 families) across all six blinding levels and three datasets (rows: QM7, Lipophilicity, Delaney). Different colors indicate model sizes within each family. Compared to the Gemini results in Figure~\ref{fig:gemini_approaches}, the OpenAI models exhibit more uniform behavior across size variants, and the two families differ markedly in which blinding regime yields peak performance.}
\label{fig:openai_approaches}
\end{center}
\end{figure*}

Figures~\ref{fig:gemini_approaches} and \ref{fig:openai_approaches} detail the performance of the Gemini 2.5 and OpenAI (GPT-4.1 and GPT-5) model families, respectively, under progressive information removal (Levels 1-6) with in-context learning and 1000 samples. By the systematic removal or transformation of information, we can estimate the influence and strength of different capabilities of the LLMs. Additionally, we can compare the underlying difficulty levels of the prediction tasks by comparing the LLM performance on the fully blinded tasks. To confirm that the transformed levels probe genuine in-context learning rather than recovery of the original value scale from a monotonic map, we additionally validated them with a non-monotonic sine transform that cannot be recovered from in-context anchors; across GPT-5, GPT-4.1, and Gemini 2.5 Pro the models retain their in-context performance under it---above both the structural-similarity and recall baselines---at every transformed level (non-monotonic transform control, Appendix). Throughout this section we emphasize that, given only two runs per configuration, only differences whose bootstrap confidence intervals separate (Appendix) should be read as robust; the smaller, few-percentage-point gaps we describe are reported as qualitative trends rather than established effects.

Gemini 2.5 Pro consistently performs best but reveals distinct sensitivities. On Lipophilicity and Delaney, performance decreases with blinding, but the drop is much steeper for Lipophilicity (72\% $\to$ 35\%) than for Delaney (96\% $\to$ 92\%). This significantly larger drop indicates substantial reliance on pre-trained knowledge (rather than verbatim memorization) for Lipophilicity, together with relatively weak in-context learning performance. In particular, a modest 5--15 percentage-point gap between transformed and untransformed values may reflect either a value-range effect or a memorization effect in which the model recognizes the original dataset values; this gap is comparable to the test-set confidence interval, so we do not over-interpret it, and---given our negative memorization finding---we tentatively attribute it to a value-range issue rather than to recall. The robustness on Delaney even under full blinding suggests both that the structure-property relationship of solubility is not too complex and that the model can learn this relationship almost completely via in-context learning.

QM7 reveals a different behavior: there is no clear decrease in performance with blinding. However, there is an oscillation between transformed and non-transformed values, but such that the transformed values are predicted with higher correlation. This might just be due to the value range of the original values (around $-1000$) being highly suboptimal for LLM processing. Also, together with the fact that the performance is not decreasing with the blinding, it suggests that the model has neither memorized the dataset nor has it learned the relationships and is using its general in-context learning capability mostly for prediction.

Gemini Flash exhibits anomalous ``interference'' behavior, often performing better at higher blinding levels on difficult datasets. On QM7, masking the property name (Agnostic) improves performance. On Lipophilicity, it peaks at Level 3. We hypothesize that for less familiar datasets, the model possesses ``wrong knowledge'' or hallucinations; blinding silences these incorrect priors, allowing the model to rely more effectively on the ground-truth in-context examples. Flash-Lite generally shows low performance, partly following Flash's trends on QM7 but struggling with larger datasets like Delaney.

While many trends are similar to the Gemini models, the OpenAI models show some distinct patterns. First, the OpenAI mini and nano variants perform much more similarly to their larger variants. This contrasts with the Gemini models, where the Flash and Flash-Lite variants show very different patterns compared to Gemini 2.5 Pro. Second, the GPT-4.1 models tend to peak with the Generic approaches and the GPT-5 models with the Specific approaches; because these peak differences are only a few percentage points and are not individually CI-separated, we report them as trends only. Taken together with the observation that both families reach similar correlation scores on the fully blinded Agnostic approaches, they are suggestive of GPT-5 combining prior knowledge and in-context samples somewhat more effectively than GPT-4.1, but this should not be read as a robust per-model claim.

The sole exception is the QM7 dataset with original values, where GPT-5 achieves much lower correlation scores. On this dataset, GPT-5 and GPT-5 mini show a similar oscillatory behavior, but with much larger amplitudes between the transformed and original values. This oscillatory behavior is also observed with Gemini 2.5 Pro and GPT-4.1, but on a much lower amplitude. This extreme sensitivity suggests that the large, negative original numbers in the QM7 dataset (approximately $-500$ to $-2500$~kcal/mol) confuse the numerical reasoning of the GPT-5 models.

\section{Discussion}

\subsection{Key Findings and Implications}

Our systematic evaluation across nine LLM variants, three datasets, and six blinding levels reveals several insights into how LLMs balance prior knowledge and in-context learning data when performing molecular property prediction.

\subsection{Memorization vs. Genuine Reasoning (Addressing RQ1)}

Our results demonstrate that LLMs rely on different capabilities when performing molecular property prediction. Crucially, the 0-shot performance on a dataset published after the models' knowledge cutoff shows that LLMs do generalize to unseen samples, consistent with their having learned a general structure-property relationship rather than retrieving specific values. The behavior on the datasets predating 2018 is very similar, indicating that predictions there, too, are not driven by direct, verbatim retrieval. We caution, however, that ruling out verbatim retrieval does not establish that the predictions arise purely from learned reasoning: approximate recall, substructure-level associations, or other statistical signals acquired during pre-training may also contribute to the observed 0-shot correlation. When given access to in-context examples, they additionally leverage in-context learning capabilities, as the increase in performance from 0-shot to 1000-shot indicates. However, a control test with IUPAC atomic weights and boiling points from Wikipedia showed that LLMs do retrieve values verbatim for numbers that are present in their training corpus.

When prediction must rely on in-context learning (enforced via blinding), performance varies substantially across datasets and model families. Lipophilicity is an example of a property that LLMs appear to recognize conceptually but struggle to learn via in-context examples, resulting in a significant performance drop from approximately 72\% to 35\% under blinding (consistent across $8/9$ models, sign-test $p=0.02$, and individually CI-separated for the strongest models; Appendix, significance analysis). Solubility, in contrast, shows that LLMs are able to learn the mapping in context, with only a minor drop from approximately 96\% to 92\%. The QM7 results even indicate no performance degradation under blinding. These differences likely reflect the underlying complexity of the structure-property relationships: Lipophilicity depends more strongly on global molecular structure, whereas solubility (Delaney) and atomization energy (QM7) are more directly associated with functional groups or atomic properties and hence can be learned more easily via in-context learning.

This progression also invites a mechanistic reading. Under full blinding the model receives only transformed structure strings and in-context examples, yet still predicts well above chance---in effect performing automated featurization followed by in-context regression (cf.\ Akyürek et al.~\yrcite{akyurek2022transformers}). A ridge regressor on a bag-of-characters representation of the SMILES string fit on 1000 examples (char-1000; Appendix, Table~S8) is never clearly outperformed by an LLM on a fully blinded approach. What the models add from their knowledge and other capabilities becomes visible only as chemical context is restored through unblinding. However, exactly this additional knowledge can also interfere with in-context learning, as we discuss in the next section. 

Taken together, these findings suggest that LLMs use their learned structure-property relationships if possible, but are also able to utilize chemical in-context learning to improve over the 0-shot setting. They even use general in-context learning in the absence of direct chemical context, although the complexity of the property to predict limits the performance of their in-context learning.

\subsection{The Double-Edged Sword of Prior Knowledge (Addressing RQ2)}

In-context learning is often framed as a process where prior knowledge facilitates new learning. Our findings refine this view: prior knowledge is a double-edged sword that can facilitate or \textit{interfere} with learning, depending on its alignment with the specific task. We observe that the peak performance is often not at blinding level 1, consistent with prior knowledge hindering in-context learning. Most individual per-level differences are within the test-set confidence interval, so we do not over-interpret them; in specific cases, however, the effect is large and statistically significant---most clearly Gemini 2.5 Flash on QM7 (discussed below), where suppressing the property prior raises the correlation by roughly $+38$ percentage points (paired-bootstrap confidence interval excluding zero). Averaged across models the property name alone is not decisive---naming it generically (Level 3) is statistically indistinguishable from naming it specifically (Level 1)---and the most robust pooled signatures of interference are the few-shot dip discussed below and the Lipophilicity collapse (Appendix, significance analysis).

Furthermore, our results show a decrease in performance when using 60 samples instead of 0 in $19$ of $27$ model$\times$dataset cases (sign-test $p=0.026$), even when the models are given reasoning time. For all models apart from Gemini 2.5 Flash and Flash-Lite, this effect is mostly mitigated by providing 1000 samples (1000-shot above 60-shot in $21/27$ cases, $p=0.003$; Appendix, significance analysis). This suggests a conflict between insufficient in-context data and the model's strong prior knowledge, where too few examples fail to override pre-existing incorrect assumptions about the task. This is mostly resolved when providing 1000 samples.

Gemini 2.5 Flash and Flash-Lite show a stronger decrease in performance, leading to a lower performance with 1000 samples than in the 0-shot setting. Interestingly, this decrease is at least partly removed upon increasing the blinding level. This indicates that these two models cannot combine their knowledge with the in-context examples. As soon as we deactivate the models' ability to use its knowledge, there is only one source of information and the models' performance increases again. On the QM7 dataset, this interaction produces particularly counterintuitive behavior: the models perform poorly without blinding, but exhibit sharp performance gains once the property name is hidden and inputs or outputs are transformed. For Gemini 2.5 Flash this gain is large and statistically significant---the QM7 correlation rises from approximately $6\%$ at Level 1 to approximately $44\%$ at Level 6, a $+38$ percentage-point improvement whose paired-bootstrap confidence interval $[+17,+54]$ excludes zero (Gemini 2.5 Flash-Lite trends the same way, $+15$ pp, within noise). This suggests that these models rely on misleading prior associations for QM7-like tasks, which are only suppressed once prior knowledge is neutralized through blinding.

In conclusion, we find evidence that prior knowledge can be beneficial, but is often detrimental, especially when the amount of in-context examples is limited. This suggests that practical LLM deployment should carefully consider whether to leverage or suppress domain priors based on prior knowledge, data and LLM capabilities.

\subsection{Implications for Benchmarking and Practice (Addressing RQ3)}

When benchmarking LLMs on molecular property prediction, standard evaluations that mix all capabilities risk measuring training data leakage or overlap specific to the benchmark rather than generalizable pure in-context learning capabilities. If the goal is to assess an LLM's ability to generalize to novel chemical spaces, benchmarks must isolate in-context learning capabilities. Our blinding protocol offers a concrete method for this: high performance that survives blinding (like Delaney) indicates robust reasoning, while performance that collapses (like Lipophilicity) indicates contamination or shallow heuristics.

For practical applications, the choice of which capabilities to leverage depends critically on the specific task requirements. General in-context learning is universally valuable, as it enables prediction on arbitrary structure-property relationships without domain-specific priors. Chemical in-context learning becomes particularly useful when the task involves molecular property prediction where the structural representation (e.g., SMILES) is familiar to the LLM. While not empirically validated in this work, we hypothesize this extends to other domains where the LLM has been exposed to the structural language during pre-training. Learned domain-specific relationships represent a double-edged sword: when the property is well-known to the LLM, leveraging this prior knowledge can substantially improve prediction performance. However, as demonstrated in our results, this same prior knowledge can diminish performance when coupled with small in-context datasets, particularly when the LLM's priors conflict with the provided examples. In such cases, strategically blinding the property name to suppress misleading priors may be beneficial.

One interesting finding is that the scale of the numbers to predict significantly affected the prediction performance at least in the case of QM7. This suggests that, in practice, rescaling the labels---as is routinely done for simple neural networks---might benefit the LLMs' predictions. 

\subsection{Limitations and Future Work}

Our study has several limitations that should be considered when interpreting results. First, we use only SMILES string representations, while many molecular property prediction methods leverage 3D coordinates or molecular graphs. This choice reflects the text-based nature of LLMs but may underestimate their potential with richer input formats. Second, our statistical power is limited by computational costs: we perform only 2 runs per configuration. We quantify the resulting uncertainty by bootstrapping over the test set (Appendix) and have softened model-to-model comparisons accordingly; nonetheless, additional runs would tighten the confidence intervals and are a natural extension. Third, a study using an LLM with an openly available training corpus would make it possible to prove to what extent LLMs memorize specific numbers and datasets, which we cannot do here. Fourth, our main blinding sweep is confined to three widely used benchmarks (ESOL, Lipophilicity, QM7) that all predate 2018. This is a deliberate choice---these datasets are the most likely to exhibit contamination---but it raises the question of whether the conclusions transfer to newer data. As a first check, our 2025 post-cutoff antiviral-potency control behaves just like the legacy sets (coincidence-level match and retention rates), suggesting the memorization conclusions are not an artifact of dataset age. But, as the memorization of the positive controls shows, there are datasets from which the LLMs do recall values verbatim. The difference is most likely their appearance or non-appearance in the training corpus, which is a property of the dataset--model pair rather than of either in isolation. On the mechanism behind this we can only speculate, since the training corpora are not accessible: what may decide it is not the age of a dataset but how its values are published, with the boiling points and atomic weights tabulated as plain text across many independent pages whereas the benchmark values circulate as separate data files. Hence, the findings in this study need to be re-evaluated for each new dataset--model pair and indicate only that these LLMs have not been trained directly on these datasets. Regarding the other findings in our study, we expect the LLMs' in-context learning capabilities to improve with the general development of LLMs and hence the knowledge conflicts to diminish with time. But this is a hypothesis that needs to be tested in future work, for example by repeating the experiments presented in this study with future LLMs and datasets. 

\section{Conclusion}

In this work, we investigated whether large language models perform genuine in-context regression for molecular property prediction or whether their apparent success is driven by memorization or prior domain knowledge and in-context learning capabilities.

Using a systematic six-level blinding framework across multiple datasets, and a memorization detector validated against a positive control of known values, we find no evidence that LLMs rely on direct, verbatim retrieval for the widely used molecular datasets predating 2018. Where prior knowledge contributes, our data are most consistent with generalized domain understanding rather than dataset-specific recall---although a verbatim-match detector cannot exclude approximate recall or substructure-level statistical associations, so we state this as the reading our evidence supports rather than as a demonstrated mechanism. For values from other sources like Wikipedia and IUPAC, by contrast, LLMs do retrieve them verbatim. 

Our results show that LLMs flexibly combine prior domain knowledge and in-context information, with this interaction both enabling and hindering performance depending on the task and the in-context sample count. In several cases, suppressing prior knowledge through blinding improves predictive accuracy, indicating that strong priors can interfere with in-context learning rather than support it. These effects vary substantially across properties and datasets, highlighting the limitations of standard benchmark evaluations. Often, giving the LLM access to 60 samples reduced the correlation compared to the 0-shot setting, while the addition of 1000 samples on average led to a significant improvement, at least for the frontier models.

Overall, our findings suggest that strong benchmark performance should not be equated with robust in-context regression. To test the pure in-context-learning abilities of LLMs, we recommend fully blinding the input data. For maximum predictive accuracy, however, whether including prior knowledge hinders performance---in the case of knowledge conflicts---or improves it depends on the specific case. Finally, rescaling the labels can further increase the predictive accuracy.

\section*{Acknowledgements}
We gratefully acknowledge the support of the Helmholtz-Gemeinschaft Deutscher Forschungszentren (HGF).

\section*{Funding}
This publication was funded via project 535656357 from Deutsche Forschungsgemeinschaft (DFG): \url{https://gepris.dfg.de/gepris/projekt/535656357}

\section*{Author contribution: CRediT}
\textbf{Matthias Busch}: Conceptualization, Methodology, Software, Formal analysis, Investigation, Writing -- Original Draft, Writing -- Review \& Editing, Visualization.
\textbf{Marius Tacke}: Writing -- Review \& Editing.
\textbf{Sviatlana V. Lamaka}: Writing -- Review \& Editing.
\textbf{Mikhail L. Zheludkevich}: Writing -- Review \& Editing.
\textbf{Christian J. Cyron}:  Writing -- Review \& Editing.
\textbf{Christian Feiler}: Supervision, Writing -- Review \& Editing.
\textbf{Roland C. Aydin}: Conceptualization, Supervision, Writing -- Review \& Editing.

\section*{Declaration of competing interests}
The authors declare no competing interests. 

\section*{Code and Data Availability}
All code, data, and pre-computed results required to reproduce the experiments and figures in this paper are publicly available at \url{https://github.com/MatthiasHBusch/BlindingLLMs}.

\bibliographystyle{plainnat}
\bibliography{references}

@article{busch2025corrosion,
  title     = {Large language models predicting the corrosion inhibition efficiency of magnesium dissolution modulators},
  author    = {Busch, Matthias and Tacke, Marius and Lamaka, Sviatlana V and Zheludkevich, Mikhail L and Linka, Kevin and Cyron, Christian J and Feiler, Christian and Aydin, Roland C},
  journal   = {Corrosion Science},
  volume    = {255},
  pages     = {113080},
  year      = {2025},
  publisher = {Elsevier}
}

@article{joe2026icl,
  title     = {Evaluating In-Context Learning in Large Language Models for Molecular Property Regression},
  author    = {Joe, Chan Young and Song, Kyungwoo and Chang, Rakwoo},
  journal   = {Journal of Computational Chemistry},
  volume    = {47},
  number    = {2},
  pages     = {e70308},
  year      = {2026},
  publisher = {Wiley Online Library}
}

@article{wu2018moleculenet,
  title     = {{MoleculeNet}: A benchmark for molecular machine learning},
  author    = {Wu, Zhenqin and Ramsundar, Bharath and Feinberg, Evan N and Gomes, Joseph and Geniesse, Caleb and Pappu, Aneesh S and Leswing, Karl and Pande, Vijay},
  journal   = {Chemical science},
  volume    = {9},
  number    = {2},
  pages     = {513--530},
  year      = {2018},
  publisher = {Royal Society of Chemistry}
}

@article{rupp2012qm7,
  title     = {Fast and accurate modeling of molecular atomization energies with machine learning},
  author    = {Rupp, Matthias and Tkatchenko, Alexandre and M{\"u}ller, Klaus-Robert and Von Lilienfeld, O Anatole},
  journal   = {Physical review letters},
  volume    = {108},
  number    = {5},
  pages     = {058301},
  year      = {2012},
  publisher = {APS}
}

@misc{huggingfaceqm7,
  author = {Hemmati, Reza},
  title  = {QM7-Dataset},
  year   = {2024},
  url    = {https://huggingface.co/datasets/HR-machine/QM7-Dataset},
  note   = {Accessed: 2026-01-14}
}

@article{delaney2004esol,
  title     = {{ESOL}: Estimating aqueous solubility directly from molecular structure},
  author    = {Delaney, John S},
  journal   = {Journal of chemical information and computer sciences},
  volume    = {44},
  number    = {3},
  pages     = {1000--1005},
  year      = {2004},
  publisher = {ACS Publications}
}

@article{bran2023chemcrow,
  title     = {Augmenting large language models with chemistry tools},
  author    = {M. Bran, Andres and Cox, Sam and Schilter, Oliver and Baldassari, Carlo and White, Andrew D and Schwaller, Philippe},
  journal   = {Nature Machine Intelligence},
  volume    = {6},
  number    = {5},
  pages     = {525--535},
  year      = {2024},
  publisher = {Nature Publishing Group UK London}
}

@inproceedings{NEURIPS2023_bbb33018,
  author    = {Guo, Taicheng and Guo, kehan and Nan, Bozhao and Liang, Zhenwen and Guo, Zhichun and Chawla, Nitesh and Wiest, Olaf and Zhang, Xiangliang},
  booktitle = {Advances in Neural Information Processing Systems},
  editor    = {A. Oh and T. Naumann and A. Globerson and K. Saenko and M. Hardt and S. Levine},
  pages     = {59662--59688},
  publisher = {Curran Associates, Inc.},
  title     = {What can Large Language Models do in chemistry? A comprehensive benchmark on eight tasks},
  volume    = {36},
  year      = {2023}
}

@article{dong2023survey,
  title     = {A survey on in-context learning},
  author    = {Dong, Qingxiu and Li, Lei and Dai, Damai and Zheng, Ce and Ma, Jingyuan and Li, Rui and Xia, Heming and Xu, Jingjing and Wu, Zhiyong and Chang, Baobao and others},
  booktitle = {Proceedings of the 2024 conference on empirical methods in natural language processing},
  pages     = {1107--1128},
  year      = {2024}
}

@inproceedings{zhou2024mystery,
  title     = {The mystery of in-context learning: A comprehensive survey on interpretation and analysis},
  author    = {Zhou, Yuxiang and Li, Jiazheng and Xiang, Yanzheng and Yan, Hanqi and Gui, Lin and He, Yulan},
  booktitle = {Proceedings of the 2024 Conference on Empirical Methods in Natural Language Processing},
  pages     = {14365--14378},
  year      = {2024}
}

@article{brown2020gpt3,
  title   = {Language models are few-shot learners},
  author  = {Brown, Tom and Mann, Benjamin and Ryder, Nick and Subbiah, Melanie and Kaplan, Jared D and Dhariwal, Prafulla and Neelakantan, Arvind and Shyam, Pranav and Sastry, Girish and Askell, Amanda and others},
  journal = {Advances in neural information processing systems},
  volume  = {33},
  pages   = {1877--1901},
  year    = {2020}
}

@inproceedings{carlini2023memorization,
  title     = {Quantifying memorization across neural language models},
  author    = {Carlini, Nicholas and Ippolito, Daphne and Jagielski, Matthew and Lee, Katherine and Tramer, Florian and Zhang, Chiyuan},
  booktitle = {The Eleventh International Conference on Learning Representations},
  year      = {2023}
}

@inproceedings{sainz2023contamination,
  title     = {NLP evaluation in trouble: On the need to measure LLM data contamination for each benchmark},
  author    = {Sainz, Oscar and Campos, Jon and Garc{\'\i}a-Ferrero, Iker and Etxaniz, Julen and de Lacalle, Oier Lopez and Agirre, Eneko},
  booktitle = {Findings of the Association for Computational Linguistics: EMNLP 2023},
  pages     = {10776--10787},
  year      = {2023}
}

@article{liu2024lost,
  title   = {Lost in the middle: How language models use long contexts},
  author  = {Liu, Nelson F and Lin, Kevin and Hewitt, John and Paranjape, Ashwin and Bevilacqua, Michele and Petroni, Fabio and Liang, Percy},
  journal = {Transactions of the association for computational linguistics},
  volume  = {12},
  pages   = {157--173},
  year    = {2024}
}

@article{weininger1988smiles,
  title     = {{SMILES}, a chemical language and information system. 1. Introduction to methodology and encoding rules},
  author    = {Weininger, David},
  journal   = {Journal of chemical information and computer sciences},
  volume    = {28},
  number    = {1},
  pages     = {31--36},
  year      = {1988},
  publisher = {ACS Publications}
}

@article{yang2019analyzing,
  title     = {Analyzing learned molecular representations for property prediction},
  author    = {Yang, Kevin and Swanson, Kyle and Jin, Wengong and Coley, Connor and Eiden, Philipp and Gao, Hua and Guzman-Perez, Angel and Hopper, Timothy and Kelley, Brian and Mathea, Miriam and others},
  journal   = {Journal of chemical information and modeling},
  volume    = {59},
  number    = {8},
  pages     = {3370--3388},
  year      = {2019},
  publisher = {ACS Publications}
}

@inproceedings{gilmer2017message,
  title        = {Neural message passing for quantum chemistry},
  author       = {Gilmer, Justin and Schoenholz, Samuel S and Riley, Patrick F and Vinyals, Oriol and Dahl, George E},
  booktitle    = {International conference on machine learning},
  pages        = {1263--1272},
  year         = {2017},
  organization = {Pmlr}
}

@article{wei2022chain,
  title   = {Chain-of-thought prompting elicits reasoning in large language models},
  author  = {Wei, Jason and Wang, Xuezhi and Schuurmans, Dale and Bosma, Maarten and Xia, Fei and Chi, Ed and Le, Quoc V and Zhou, Denny and others},
  journal = {Advances in neural information processing systems},
  volume  = {35},
  pages   = {24824--24837},
  year    = {2022}
}

@article{jablonka2024leveraging,
  title     = {Leveraging large language models for predictive chemistry},
  author    = {Jablonka, Kevin Maik and Schwaller, Philippe and Ortega-Guerrero, Andres and Smit, Berend},
  journal   = {Nature Machine Intelligence},
  volume    = {6},
  number    = {2},
  pages     = {161--169},
  year      = {2024},
  publisher = {Nature Publishing Group UK London}
}

@misc{zhang2024chemllm,
  title         = {{ChemLLM}: A chemical large language model},
  author        = {Zhang, Di and Liu, Wei and Tan, Qian and Chen, Jingdan and Yan, Hang and Yan, Yuliang and Li, Jiatong and Huang, Weiran and Yue, Xiangyu and Ouyang, Wanli and others},
  year          = {2024},
  eprint        = {2402.06852},
  archiveprefix = {arXiv},
  primaryclass  = {cs.CL},
  url           = {https://arxiv.org/abs/2402.06852}
}

@misc{akyurek2022transformers,
  title         = {What learning algorithm is in-context learning? investigations with linear models},
  author        = {Aky{\"u}rek, Ekin and Schuurmans, Dale and Andreas, Jacob and Ma, Tengyu and Zhou, Denny},
  year          = {2022},
  eprint        = {2211.15661},
  archiveprefix = {arXiv},
  primaryclass  = {cs.LG},
  url           = {https://arxiv.org/abs/2211.15661}
}

@misc{cheng2025survey,
  title         = {A survey on data contamination for large language models},
  author        = {Cheng, Yuxing and Chang, Yi and Wu, Yuan},
  year          = {2025},
  eprint        = {2502.14425},
  archiveprefix = {arXiv},
  primaryclass  = {cs.CL},
  url           = {https://arxiv.org/abs/2502.14425}
}

@inproceedings{chen2025benchmarking,
  title     = {Benchmarking large language models under data contamination: A survey from static to dynamic evaluation},
  author    = {Chen, Simin and Chen, Yiming and Li, Zexin and Jiang, Yifan and Wan, Zhongwei and He, Yixin and Ran, Dezhi and Gu, Tianle and Li, Haizhou and Xie, Tao and others},
  booktitle = {Proceedings of the 2025 Conference on Empirical Methods in Natural Language Processing},
  pages     = {10091--10109},
  year      = {2025}
}

@misc{bordt2025forget,
  title         = {How much can we forget about data contamination?},
  author        = {Bordt, Sebastian and Srinivas, Suraj and Boreiko, Valentyn and von Luxburg, Ulrike},
  year          = {2024},
  eprint        = {2410.03249},
  archiveprefix = {arXiv},
  primaryclass  = {cs.LG},
  url           = {https://arxiv.org/abs/2410.03249}
}

@misc{asap_polaris_openadmet_2025,
  author    = {Scheen, Jenke and MacDermott-Opeskin, Hugo and Chodera, John and
               Griffen, Edward and von Delft, Frank and Aschenbrenner, Jasmin Cara and
               Fearon, Daren and Balcomb, Blake and Marples, Peter and
               Tomlinson, Charles W. E. and Lithgo, Ryan and Winokan, Max and
               Cousins, David and Stacey, Jessica and Barr, Haim},
  title     = {{ASAP-Polaris-OpenADMET} Blind Challenge Data},
  year      = {2025},
  publisher = {Zenodo},
  doi       = {10.5281/zenodo.15582067},
  url       = {https://doi.org/10.5281/zenodo.15582067},
  note      = {Version v2, dataset}
}

\appendix

\newpage

\section{Memorization Analysis: Full Results}
\label{SI:memorization}

\begin{figure}[htbp]
    \centering
    \begin{subfigure}{0.48\textwidth}
        \includegraphics[width=\linewidth]{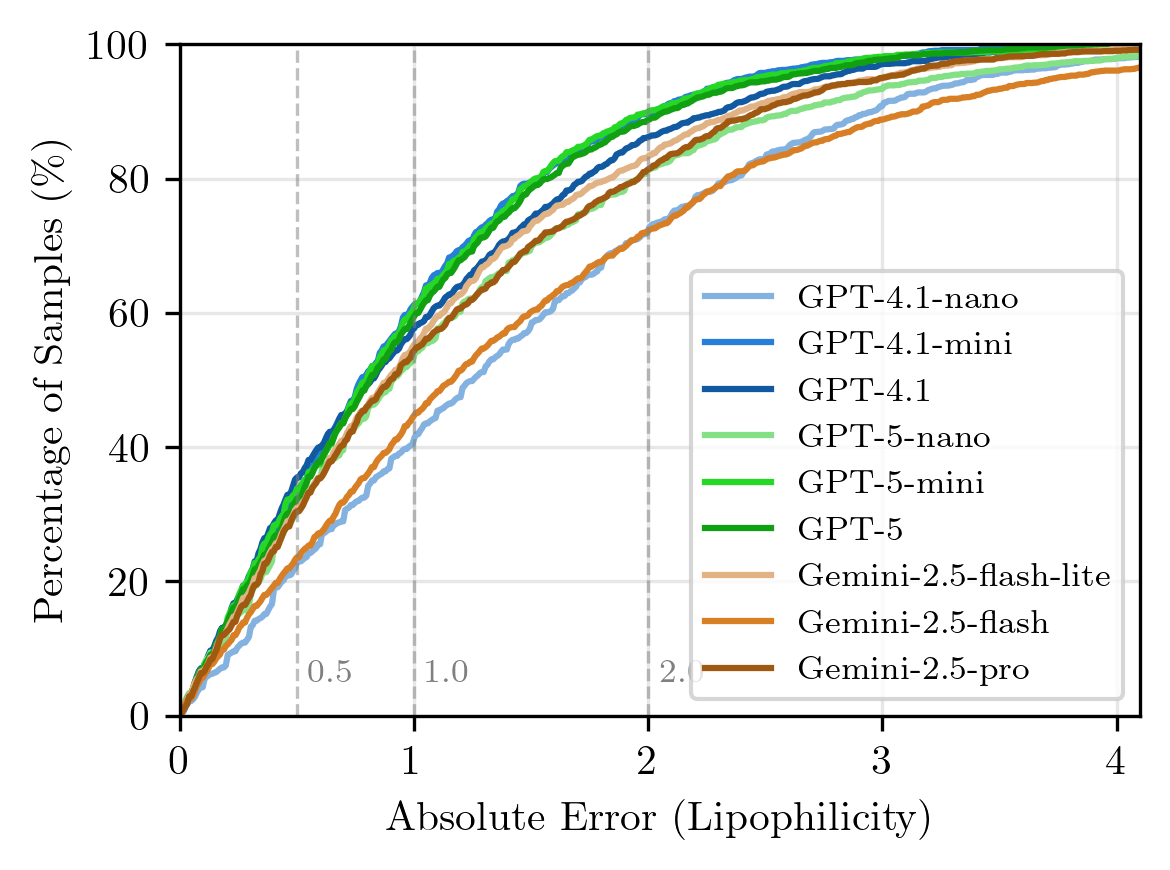}
        \caption{Lipophilicity}
    \end{subfigure}
    \hfill
    \begin{subfigure}{0.48\textwidth}
        \includegraphics[width=\linewidth]{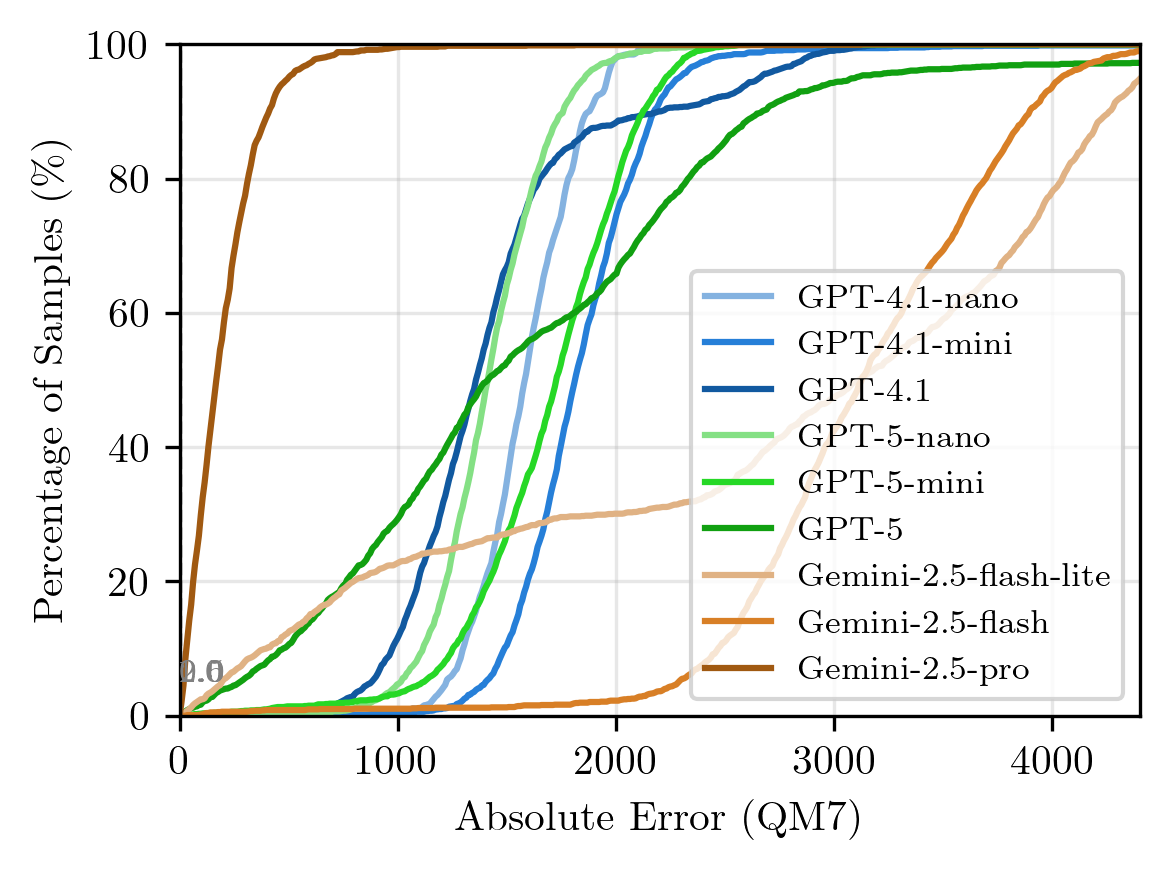}
        \caption{QM7}
    \end{subfigure}
    \caption{Cumulative error distributions for Lipophilicity (a) and QM7 (b) datasets. Some observations: The cumulative error on the Lipophilicity dataset has ragged edges, appearing discontinuous. This is because the dataset contains many low-precision values, which group many error values together. The predictions for the QM7 dataset, however, are highly inaccurate; common errors include predictions at the wrong scale (around 100) and with the wrong sign. Since the original QM7 values lie in the range $-500$ to $-2500$, this explains the errors. Typically, these LLMs make all predictions at the wrong scale, which is a sign of misaligned priors. This might also be part of the information conflict between prior knowledge and in-context samples that we assume for Gemini 2.5 Flash and Flash-Lite.}
    \label{SI-fig:error_distributions}
\end{figure}

\begin{table}[htbp]
\centering
\caption{Per-model memorization diagnostics on the three legacy benchmarks (0-shot, five
repetitions). For each dataset: \emph{3s} = three-significant-digit exact-match rate (\%, over
ground-truth values with $\geq$3 significant figures); $R_{21}=m_2/m_1$ and $R_{32}=m_3/m_2$ =
precision-scaling retention (\%; coincidence floor $\approx10\%$, genuine recall $\approx100\%$);
$r$ = Pearson correlation of the per-molecule mean prediction. ``--'' marks cells with no
qualifying lower-precision match. Per-model retention cells (especially $R_{32}$, and most of
QM7) rest on few matches and are noisy; the pooled row (count-pooled rates; median $r$) is the
robust summary.}
\label{SI-tab:mem_benchmarks}
\small
\resizebox{\textwidth}{!}{%
\begin{tabular}{l|rrrr|rrrr|rrrr}
\toprule
 & \multicolumn{4}{c|}{\textbf{Delaney (ESOL)}} & \multicolumn{4}{c|}{\textbf{Lipophilicity}} & \multicolumn{4}{c}{\textbf{QM7}} \\
\textbf{Model} & 3s & $R_{21}$ & $R_{32}$ & $r$ & 3s & $R_{21}$ & $R_{32}$ & $r$ & 3s & $R_{21}$ & $R_{32}$ & $r$ \\
\midrule
gemini-2.5-pro        & 0.55 & 12.3 & 14.6 &  0.89 & 0.50 & 11.2 & 14.0 & 0.47 & 1.22 & 20.6 &  7.4 &  0.55 \\
gemini-2.5-flash      & 0.64 & 11.2 & 15.1 &  0.88 & 0.36 & 11.2 & 14.9 & 0.43 & 0.12 & 16.1 & 14.6 & -0.31 \\
gemini-2.5-flash-lite & 0.30 & 10.2 & 11.0 &  0.68 & 0.29 & 11.3 &  7.1 & 0.43 & 0.06 &  5.9 &  8.1 & -0.01 \\
gpt-4.1               & 0.39 & 10.4 & 14.3 &  0.86 & 0.57 & 11.3 & 12.8 & 0.44 & 0.00 &  --  &  --  &  0.11 \\
gpt-4.1-mini          & 0.28 & 11.2 &  8.5 &  0.78 & 0.25 & 11.8 &  5.2 & 0.36 & 0.00 &  --  &  --  & -0.04 \\
gpt-4.1-nano          & 0.14 &  9.7 &  9.0 &  0.41 & 0.00 & 10.0 &  --  & 0.29 & 0.00 &  --  &  --  & -0.01 \\
gpt-5                 & 0.37 & 11.6 & 13.3 &  0.85 & 0.61 & 11.9 & 14.2 & 0.38 & 0.04 & 12.9 &  4.3 &  0.01 \\
gpt-5-mini            & 0.18 &  9.9 &  7.5 & -0.02 & 0.36 & 12.3 & 11.9 & 0.39 & 0.00 & 13.6 &  0.0 &  0.02 \\
gpt-5-nano            & 0.14 &  9.4 & 16.7 &  0.63 & 0.00 & 13.0 &  0.0 & 0.25 & 0.00 &  0.0 &  0.0 &  0.03 \\
\midrule
\textbf{Pooled}       & 0.33 & 10.8 & 12.2 &  0.78 & 0.33 & 11.6 & 11.0 & 0.39 & 0.16 & 17.8 &  7.6 &  0.01 \\
\bottomrule
\end{tabular}%
}
\end{table}

\begin{table}[htbp]
\centering
\caption{Per-model memorization diagnostics on the three control datasets (0-shot, five
repetitions): the post-cutoff antiviral potency \emph{negative} control (provably unseen) and
the atomic-weight and boiling-point \emph{positive} controls (verbatim recall expected).
Columns as in Table~\ref{SI-tab:mem_benchmarks}. The detector returns coincidence-floor
retention on the unseen antiviral set---just like the legacy benchmarks---but near-total
retention on both positive-control sets, with match rates and retention scaling with model
capability. ``--'' marks cells with no qualifying lower-precision match (low-count, noisy).}
\label{SI-tab:mem_controls}
\small
\resizebox{\textwidth}{!}{%
\begin{tabular}{l|rrrr|rrrr|rrrr}
\toprule
 & \multicolumn{4}{c|}{\textbf{Antiviral potency (post-cutoff)}} & \multicolumn{4}{c|}{\textbf{Atomic weights (memorized)}} & \multicolumn{4}{c}{\textbf{Boiling points (memorized)}} \\
\textbf{Model} & 3s & $R_{21}$ & $R_{32}$ & $r$ & 3s & $R_{21}$ & $R_{32}$ & $r$ & 3s & $R_{21}$ & $R_{32}$ & $r$ \\
\midrule
gemini-2.5-pro        & 0.33 & 10.2 & 10.3 & 0.41 & 100.0 & 100.0 & 100.0 & 1.00 & 75.1 & 93.1 & 92.5 & 0.91 \\
gemini-2.5-flash      & 0.02 &  7.1 &  0.0 & 0.17 &  95.6 & 100.0 &  97.5 & 1.00 & 60.0 & 90.8 & 83.7 & 0.95 \\
gemini-2.5-flash-lite & 0.26 & 12.3 & 16.2 & 0.21 &  91.7 & 100.0 &  97.4 & 0.95 & 37.1 & 72.8 & 82.9 & 0.80 \\
gpt-4.1               & 0.04 & 10.0 &  8.7 & 0.69 &  93.2 & 100.0 &  97.4 & 0.70 & 43.9 & 79.5 & 81.7 & 0.91 \\
gpt-4.1-mini          & 0.00 & 10.6 &  0.0 & 0.70 &  97.6 & 100.0 &  97.6 & 1.00 & 44.4 & 86.3 & 87.8 & 0.88 \\
gpt-4.1-nano          & 0.00 &  8.8 &  --  & 0.03 &  97.6 & 100.0 &  97.5 & 1.00 & 15.6 & 55.8 & 67.5 & 0.66 \\
gpt-5                 & 0.24 &  8.6 & 11.6 & 0.46 &  94.1 &  99.0 &  97.5 & 0.67 & 46.8 & 87.9 & 79.6 & 0.81 \\
gpt-5-mini            & 0.15 &  7.0 &  6.9 & 0.37 &  97.1 & 100.0 &  97.5 & 1.00 & 49.8 & 85.5 & 87.4 & 0.90 \\
gpt-5-nano            & 0.09 & 10.4 & 40.0 & 0.32 &  96.1 & 100.0 &  97.0 & 0.99 &  6.8 & 62.7 & 54.2 & 0.59 \\
\midrule
\textbf{Pooled}       & 0.13 &  9.3 & 11.3 & 0.37 &  95.9 &  99.9 &  97.7 & 1.00 & 42.2 & 81.7 & 83.8 & 0.88 \\
\bottomrule
\end{tabular}%
}
\end{table}

Tables~\ref{SI-tab:mem_benchmarks} and \ref{SI-tab:mem_controls} show a sharp split. On the
positive control the detector fires unambiguously: three-significant-digit match rates of
$\sim$96\% (atomic weights) and $\sim$42\% pooled (boiling points, up to $75\%$ for the
strongest model), with retention near-total at both precision steps ($R_{21},R_{32}\gtrsim80\%$).
On the three legacy benchmarks and the post-cutoff antiviral control the picture is the
opposite: match rates are near zero ($\leq$1.2\% in every cell, pooled $0.13$--$0.33\%$)
and---decisively---retention sits at the $\sim$10\% coincidence floor at \emph{both} precision
steps ($R_{21}=9$--$18\%$, $R_{32}=8$--$12\%$ pooled). Because the antiviral set is provably
outside the training corpus and yet behaves identically to the legacy benchmarks, the legacy
near-zero rates are true negatives, not an insensitive detector failing to find contamination
that is present.

Two further points. First, QM7's pooled $R_{21}$ (17.8\%) sits mildly above the floor, but this
is driven almost entirely by Gemini~2.5~Pro ($R_{21}=20.6\%$ over the many like-signed,
large-magnitude atomization energies) and does \emph{not} carry to the third digit
($R_{32}=7.4\%$), the opposite of what recall would produce; it reflects scale clustering
(shown below, Figure~\ref{SI-fig:qm7digit}), not retrieval. Second, the Pearson columns confirm that correlation and memorization are distinct
axes: Delaney is well predicted (median $r=0.78$) yet at chance for recall, QM7 shows
essentially no 0-shot correlation (median $r=0.01$), and the antiviral control reaches a
moderate $r=0.37$ with a near-zero match rate and a negative $R^2$ (correct ordering but wrong
absolute scale). Only the positive-control sets combine high correlation \emph{and} high
retention---the joint signature of recall---which no property benchmark displays.

\begin{figure}[htbp]
\centering
\includegraphics[width=0.85\linewidth]{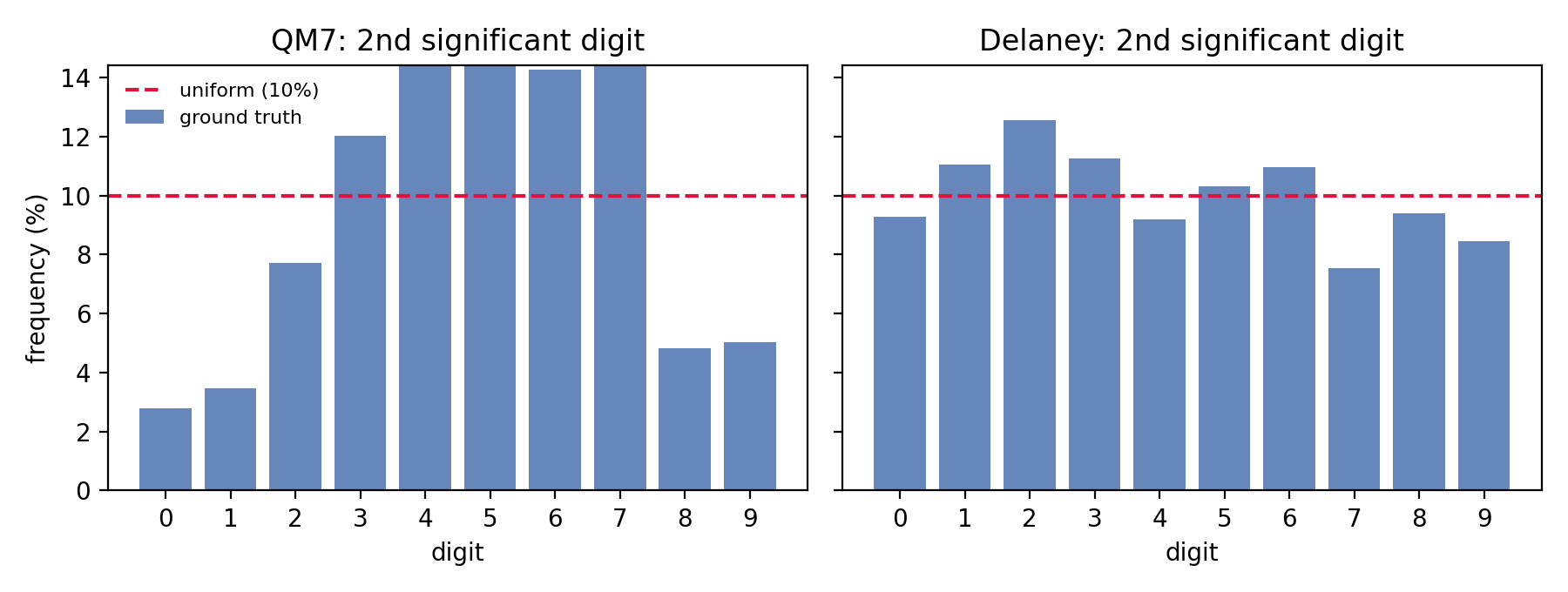}
\caption{Distribution of the second significant digit of the ground-truth values for QM7 (left)
and Delaney (right), against the uniform $10\%$ expectation (dashed red). QM7's second digit is
visibly clustered (collision probability $C_2=13.1\%$; $\chi^2_9=2116$, $p<10^{-300}$), which
raises the coincidence floor for two-digit retention above $10\%$ and accounts for its elevated
$R_{21}$ without any value recall; Delaney's second digit is essentially uniform ($C_2=10.2\%$)
and its $R_{21}$ sits at the $10\%$ floor. With the permutation test (text), this establishes the
QM7 elevation as a property of the value distribution, not of memorization.}
\label{SI-fig:qm7digit}
\end{figure}

\begin{table}[htbp]
\centering
\caption{Absolute three-significant-digit exact-match counts per model and dataset (0-shot,
five repetitions): the raw ``hits'' underlying the match rates in
Tables~\ref{SI-tab:mem_benchmarks} and \ref{SI-tab:mem_controls}. The denominator $n$ (number of
predictions over ground-truth values with $\geq$3 significant figures) is fixed per dataset and
given in the header.}
\label{SI-tab:mem_absolute}
\small
\resizebox{0.8\textwidth}{!}{%
\begin{tabular}{l|rrr|rrr}
\toprule
 & \multicolumn{3}{c|}{\textbf{Benchmarks}} & \multicolumn{3}{c}{\textbf{Controls}} \\
\textbf{Model} & \shortstack{Delaney\\{\footnotesize $n{=}4360$}} & \shortstack{Lipo.\\{\footnotesize $n{=}2800$}} & \shortstack{QM7\\{\footnotesize $n{=}4995$}} & \shortstack{Antiviral\\{\footnotesize $n{=}4570$}} & \shortstack{At.\ wt.\\{\footnotesize $n{=}205$}} & \shortstack{Boil.\ pt.\\{\footnotesize $n{=}205$}} \\
\midrule
gemini-2.5-pro        & 24 & 14 & 61 & 15 & 205 & 154 \\
gemini-2.5-flash      & 28 & 10 &  6 &  1 & 196 & 123 \\
gemini-2.5-flash-lite & 13 &  8 &  3 & 12 & 188 &  76 \\
gpt-4.1               & 17 & 16 &  0 &  2 & 191 &  90 \\
gpt-4.1-mini          & 12 &  7 &  0 &  0 & 200 &  91 \\
gpt-4.1-nano          &  6 &  0 &  0 &  0 & 200 &  32 \\
gpt-5                 & 16 & 17 &  2 & 11 & 193 &  96 \\
gpt-5-mini            &  8 & 10 &  0 &  7 & 199 & 102 \\
gpt-5-nano            &  6 &  0 &  0 &  4 & 197 &  14 \\
\midrule
\textbf{Total}        & 130 & 82 & 72 & 52 & 1769 & 778 \\
\bottomrule
\end{tabular}%
}
\end{table}

Table~\ref{SI-tab:mem_absolute} gives the underlying absolute three-significant-digit hit counts
per model and dataset. We emphasise scope: this is not a \emph{proof} that the benchmarks are
absent from training (that would require access to the undisclosed corpora), but convergent
evidence---near-zero matches, coincidence-floor retention at \emph{both} precision steps, and no
model-size scaling, against a positive control where all three signals are present---that the
models were most likely not trained on these specific benchmark tables. General conceptual
knowledge of widely tabulated properties remains legitimate prior knowledge, whose contribution
the blinding protocol is designed to isolate (main text).

\section{Blinding the Positive Control}
\label{SI:blindcontrol}

The blinding levels in the main text are comparative: they tell us how much performance a model loses when information is withheld, but not what a level does to a value the model demonstrably \emph{knows}. To calibrate this we applied the blinding levels to both positive-control sets, whose values the models reproduce verbatim at 0-shot (Table~\ref{SI-tab:mem_controls}), and asked which of the three manipulations our levels combine actually suppresses that reproduction.

\subsection{Design}

Two sweeps were run, each with 3-fold cross validation, ten repetitions, and the nine models of the main study:

\begin{enumerate}
  \item \textbf{Boiling points} (primary) --- 45 English-Wikipedia boiling points, all six levels; 30 in-context examples and 15 test compounds per fold, hence 450 predictions per model and level.
  \item \textbf{Atomic weights} (replication) --- 41 IUPAC atomic weights, levels 1--4; 28 in-context examples, 39 of the 41 elements tested per repetition, hence 390 predictions per model and level.
\end{enumerate}

Both sweeps use \textbf{single-step prompts}: the in-context examples, the target species, and the instruction to report a single numerical value. Level by level they carry the same information as the corresponding conditions of Figures~3 and~4, so the two sets are measured on the same footing and can be read side by side in Table~\ref{SI-tab:blindcontrol}. This is not, however, the benchmark prompt sequence, which instructs the model to identify similar molecules, compute a weighted average of their values, and report that average. A model that follows this instruction cannot emit an exactly recalled number even when it has one, so the exact-match detector systematically under-reports recall under that schema. In a parallel sweep under that schema the LLMs still recalled values verbatim, but at a markedly lower rate.

These experiments are based on the two positive-control datasets, which consist of 41 and 45 samples, so a 1000-shot setting as in the main experiments is not possible. We use 3-fold cross validation with 28--30 in-context examples and the remainder as the test set. Because of the small test set, we increase the number of repetitions to ten. Also, for the case of the atomic weights, the Agnostic blinding levels (5 and 6) are not defined, because each element occurs only once in the dataset, so no mapping from the structure string to the value can be learned from the in-context examples.

\begin{table}[htbp]
\centering
\caption{Blinding the two positive controls, pooled over the nine models and ten repetitions. Both sweeps use single-step prompts and 3-fold cross validation (see text): 30 in-context examples and 45 test compounds for the boiling points, 28 examples and 39 test elements for the atomic weights, so each cell rests on about 3500 predictions. $r$ = mean Pearson $|r|\times100$ against the level's own target; $m_3$ = three-significant-digit exact-match rate against that target (\%, counts pooled over models and repetitions); $R_{21}=m_2/m_1$ = precision-scaling retention in the convention of Table~\ref{SI-tab:mem_benchmarks} (coincidence floor $\approx$10\%, genuine recall $\approx$100\%). The target is the tabulated value at the untransformed levels 1/3/5 and the rescaled value at the transformed levels 2/4/6, where a match therefore also requires the affine map to be reconstructed from the in-context anchors. The Agnostic levels are not defined for the atomic weights (see text).}
\label{SI-tab:blindcontrol}
\small
\begin{tabular}{l|rrr|rrr}
\toprule
& \multicolumn{3}{c|}{\textbf{Boiling points} (45 compounds)} & \multicolumn{3}{c}{\textbf{Atomic weights} (41 elements)} \\
\textbf{Blinding level} & $r$ & $m_3$ & $R_{21}$ & $r$ & $m_3$ & $R_{21}$ \\
\midrule
1 Specific, orig. & 97 & 61.1 & 87 & 100 & 97.6 & 100 \\
2 Specific, transf. & 60 & 10.2 & 52 & 76 & 40.5 & 77 \\
3 Generic, orig. & 85 & 51.4 & 83 & 99 & 97.0 & 100 \\
4 Generic, transf. & 40 & 2.1 & 27 & 63 & 3.0 & 31 \\
5 Agnostic, orig. & 41 & 0.4 & 18 & -- & -- & -- \\
6 Agnostic, transf. & 24 & 0.3 & 13 & -- & -- & -- \\
\bottomrule
\end{tabular}
\end{table}

\begin{table}[htbp]
\centering
\caption{Per-model precision-scaling retention in the primary sweep (45 Wikipedia boiling points, single-step prompts): $R_{21}=m_2/m_1$ (\%) with a 95\% confidence interval, per model and blinding level. Retention is reported in preference to the exact-match rate because it conditions on the first significant digit already being correct and asks only whether the next digit follows; it is therefore insensitive to how accurately a model predicts the property, which at the unblinded levels is high enough that a three-digit hit need not indicate retrieval. Intervals are cluster bootstraps resampling the 45 compounds (4000 resamples) rather than the individual queries: the ten repetitions run over the same species, so they reduce noise in the models' answers but add no information about the sample of compounds, and an interval over the pooled events would be far too narrow. The last row is the floor, the probability that two distinct values of the set share their second significant digit, estimated by the U-statistic $\sum_d n_d(n_d-1)/n(n-1)$; at $n=45$ the plug-in estimator $\sum_d p_d^2$ is biased upward and would return $12.5\%$. $^{*}$ marks cells whose interval lies entirely above the floor, $^{\dagger}$ entirely below. The last column repeats each model's 0-shot retention on the same set from Table~\ref{SI-tab:mem_controls}; the Level-3 column correlates with it at $r=+0.83$ (permutation $p=0.021$, $n=9$).}
\label{SI-tab:blindcontrol_permodel}
\small
\resizebox{\textwidth}{!}{%
\begin{tabular}{l|cccccc|r}
\toprule
 & \multicolumn{6}{c|}{\textbf{Blinding level} ($R_{21}$, \%, [95\% CI])} & \textbf{0-shot} \\
\textbf{Model} & 1 Spec. & 2 Spec.-T & 3 Gen. & 4 Gen.-T & 5 Agn. & 6 Agn.-T & $R_{21}$ \\
\midrule
Gemini 2.5 Pro & 91$^{*}$ {\scriptsize[82,98]} & 74$^{*}$ {\scriptsize[66,82]} & 93$^{*}$ {\scriptsize[85,98]} & 30$^{*}$ {\scriptsize[21,39]} & 27$^{*}$ {\scriptsize[13,41]} & 12 {\scriptsize[3,22]} & 93 \\
Gemini 2.5 Flash & 85$^{*}$ {\scriptsize[76,93]} & 79$^{*}$ {\scriptsize[70,87]} & 81$^{*}$ {\scriptsize[73,89]} & 56$^{*}$ {\scriptsize[44,68]} & 21 {\scriptsize[6,38]} & 19 {\scriptsize[4,36]} & 91 \\
Gemini 2.5 Flash-Lite & 89$^{*}$ {\scriptsize[81,95]} & 39$^{*}$ {\scriptsize[28,50]} & 88$^{*}$ {\scriptsize[80,94]} & 15 {\scriptsize[5,31]} & 20 {\scriptsize[7,36]} & 16 {\scriptsize[0,33]} & 73 \\
GPT-5 & 92$^{*}$ {\scriptsize[86,97]} & 31$^{*}$ {\scriptsize[20,42]} & 92$^{*}$ {\scriptsize[85,98]} & 27$^{*}$ {\scriptsize[13,42]} & 15 {\scriptsize[6,26]} & 20 {\scriptsize[8,33]} & 88 \\
GPT-5 mini & 93$^{*}$ {\scriptsize[86,99]} & 0$^{\dagger}$ {\scriptsize[0,0]} & 85$^{*}$ {\scriptsize[77,92]} & 5 {\scriptsize[0,11]} & 6 {\scriptsize[1,12]} & 16 {\scriptsize[6,28]} & 86 \\
GPT-5 nano & 66$^{*}$ {\scriptsize[52,77]} & 9 {\scriptsize[1,19]} & 11 {\scriptsize[5,18]} & 12 {\scriptsize[6,19]} & 13 {\scriptsize[3,25]} & 10 {\scriptsize[3,19]} & 63 \\
GPT-4.1 & 94$^{*}$ {\scriptsize[88,99]} & 50$^{*}$ {\scriptsize[39,62]} & 93$^{*}$ {\scriptsize[86,98]} & 19 {\scriptsize[7,33]} & 19 {\scriptsize[6,35]} & 10 {\scriptsize[2,19]} & 80 \\
GPT-4.1 mini & 88$^{*}$ {\scriptsize[81,94]} & 13 {\scriptsize[6,21]} & 80$^{*}$ {\scriptsize[71,88]} & 13 {\scriptsize[5,24]} & 18 {\scriptsize[7,31]} & 4$^{\dagger}$ {\scriptsize[0,8]} & 86 \\
GPT-4.1 nano & 78$^{*}$ {\scriptsize[67,87]} & 0$^{\dagger}$ {\scriptsize[0,0]} & 27$^{*}$ {\scriptsize[18,36]} & 5$^{\dagger}$ {\scriptsize[0,10]} & 9 {\scriptsize[3,17]} & 2$^{\dagger}$ {\scriptsize[0,6]} & 56 \\
\midrule
\textbf{Pooled} & 87$^{*}$ {\scriptsize[81,92]} & 52$^{*}$ {\scriptsize[45,58]} & 83$^{*}$ {\scriptsize[77,88]} & 27$^{*}$ {\scriptsize[20,35]} & 18 {\scriptsize[10,27]} & 13 {\scriptsize[7,20]} & \\
\textit{floor $C_2$} & 10.5 & 10.3 & 10.5 & 10.3 & 10.5 & 10.3 & \\
\bottomrule
\end{tabular}%
}
\end{table}

\subsection{Which Manipulation Suppresses Verbatim Recall}

Masking the property name has a small effect (Level~1 to Level~3): retention falls only from $R_{21}=87\%$ to $83\%$, a twentieth of the distance down to the $\approx$10\% floor, and it stays far above that floor for eight of the nine models (Table~\ref{SI-tab:blindcontrol_permodel}). A model that knows these values evidently infers the withheld property from the untransformed in-context values and retrieves the answer regardless, which gives a mechanistic reading of the main-text observation that Level~3 is statistically indistinguishable from Level~1.

At Level~2 the property is still named and only the labels are rescaled. Retention drops substantially, from $R_{21}=87\%$ to $52\%$ on the boiling points and from $100\%$ to $77\%$ on the atomic weights, and combining the rescaling with the generic naming (Level~4) drops it further, to $27\%$ and $31\%$. The per-model breakdown shows that these pooled figures average two different regimes. For the weaker models the transformation alone already suffices---GPT-5~mini and GPT-4.1~nano fall to $R_{21}=0$ at Level~2, and at Level~4 five of the nine models are no longer distinguishable from the floor---whereas the strongest models recover the mapping with little difficulty: Gemini~2.5~Pro and Gemini~2.5~Flash still retain $74\%$ and $79\%$ at Level~2, and Gemini~2.5~Flash $56\%$ even at Level~4. What these models are doing is recalling the tabulated value and reconstructing the affine map from the in-context anchors; the map is monotone, and its recoverability is precisely the objection that motivated the non-monotonic transform control of Section~\ref{SI:nonmono}. A monotone value transformation is therefore a partial defence whose effectiveness depends on the capability of the model it is applied to, not a general one.

Transforming the structure representation (Levels~5 and 6, defined for the boiling points only) brings retention down to the floor: $R_{21}=18\%$ $[10,27]$ at Level~5 and $13\%$ $[7,20]$ at Level~6, both intervals containing the $\approx$10\% coincidence level, against $87\%$ at Level~1. At Level~6 no model lies significantly above the floor and two lie significantly below it; at Level~5 only Gemini~2.5~Pro does. We therefore read this as near-complete rather than demonstrably complete suppression. The pooled point estimates sit somewhat above the floor, which is consistent with a small residual leakage through the occasional exactly reproduced number, but with 45 compounds the intervals cannot separate that residual from coincidence. What the data do establish is that the great majority of values are no longer recalled, which is what the blinding is meant to achieve. However, it remains unclear how exactly this translates to cases without memorization, as in the main experiments.



\section{Non-Monotonic Transform Control}

\label{SI:nonmono}

\begin{table}[htbp]
\centering
\caption{Non-monotonic (sine) transform control: affine vs.\ sine $|r|\times100$ for GPT-5,
GPT-4.1, and Gemini~2.5~Pro on Delaney, 1000-shot, at the three transformed blinding levels
(same 150-molecule test set). The sine target is essentially uncorrelated with the true value,
so the recall/prior-knowledge ceiling is $\approx$19; the kNN-Tanimoto structural ceiling
(structure $\rightarrow$ sine target, hence level-independent) is $49$ under the sine transform
versus $81$ under the affine transform. Every model at every level stays well above both ceilings.}
\label{SI-tab:nonmono}
\small
\begin{tabular}{l|cc|cc|cc}
\toprule
 & \multicolumn{2}{c|}{\textbf{GPT-5}} & \multicolumn{2}{c|}{\textbf{GPT-4.1}} & \multicolumn{2}{c}{\textbf{Gemini 2.5 Pro}} \\
\textbf{Level} & affine & sine & affine & sine & affine & sine \\
\midrule
L2 Specific-T & 91 & 74 & 90 & 70 & 96 & 73 \\
L4 Generic-T  & 68 & 73 & 88 & 69 & 96 & 75 \\
L6 Agnostic-T & 86 & 69 & 82 & 63 & 92 & 70 \\
\bottomrule
\end{tabular}
\end{table}

\begin{figure}[htbp]
\centering
\includegraphics[width=0.62\linewidth]{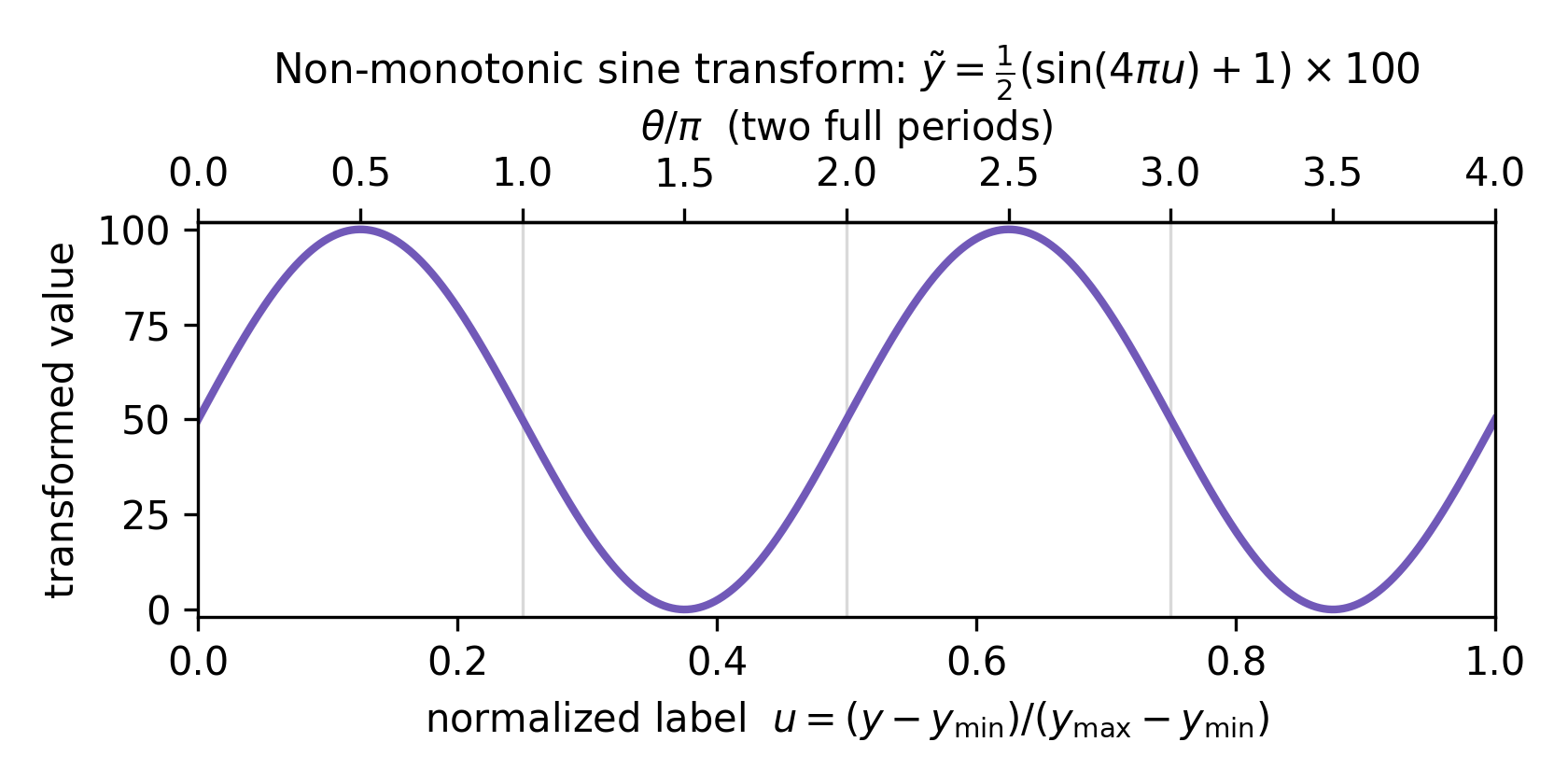}\\[0.6em]
\begin{subfigure}{0.49\textwidth}\includegraphics[width=\linewidth]{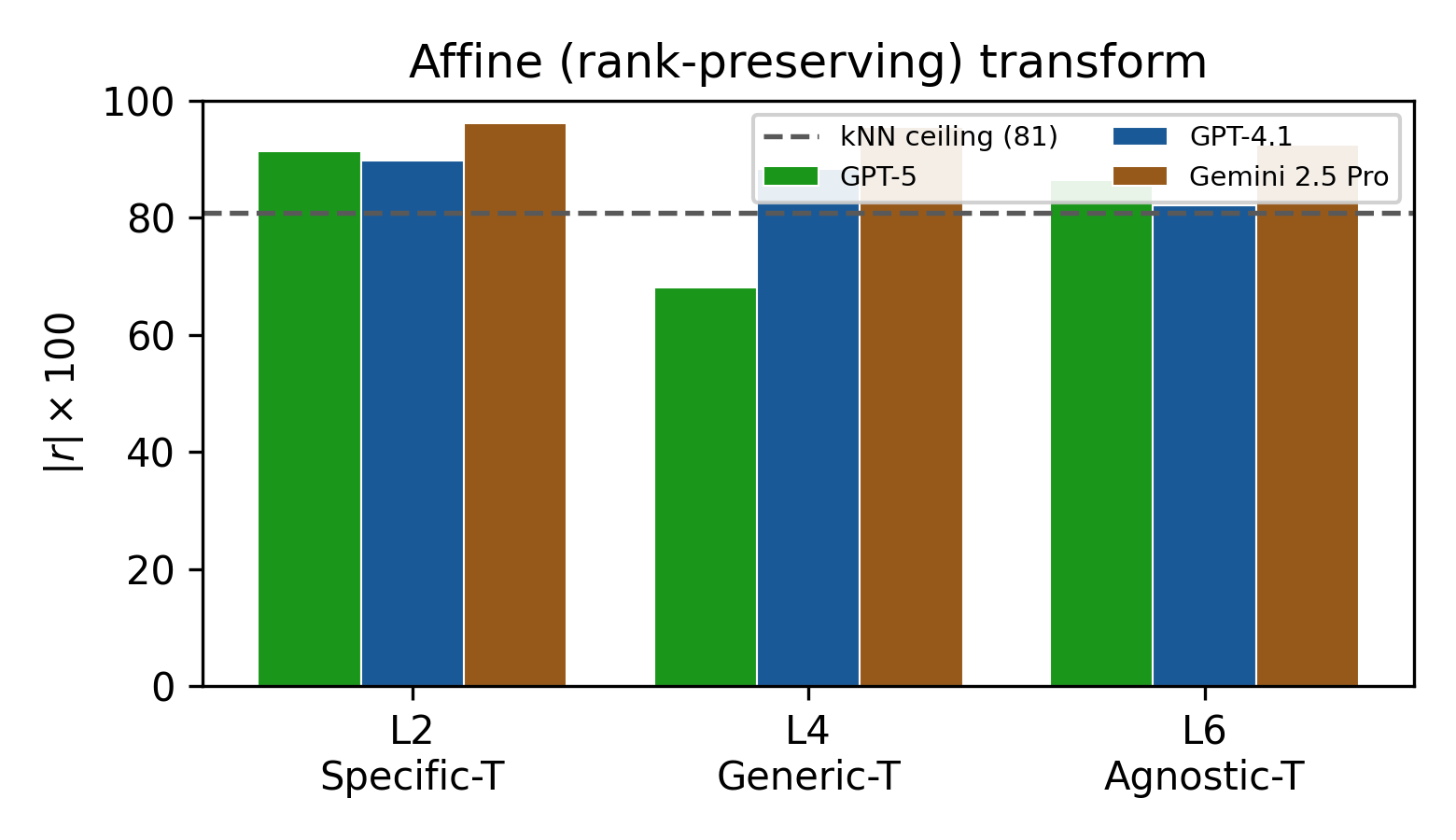}\caption{}\end{subfigure}
\hfill
\begin{subfigure}{0.49\textwidth}\includegraphics[width=\linewidth]{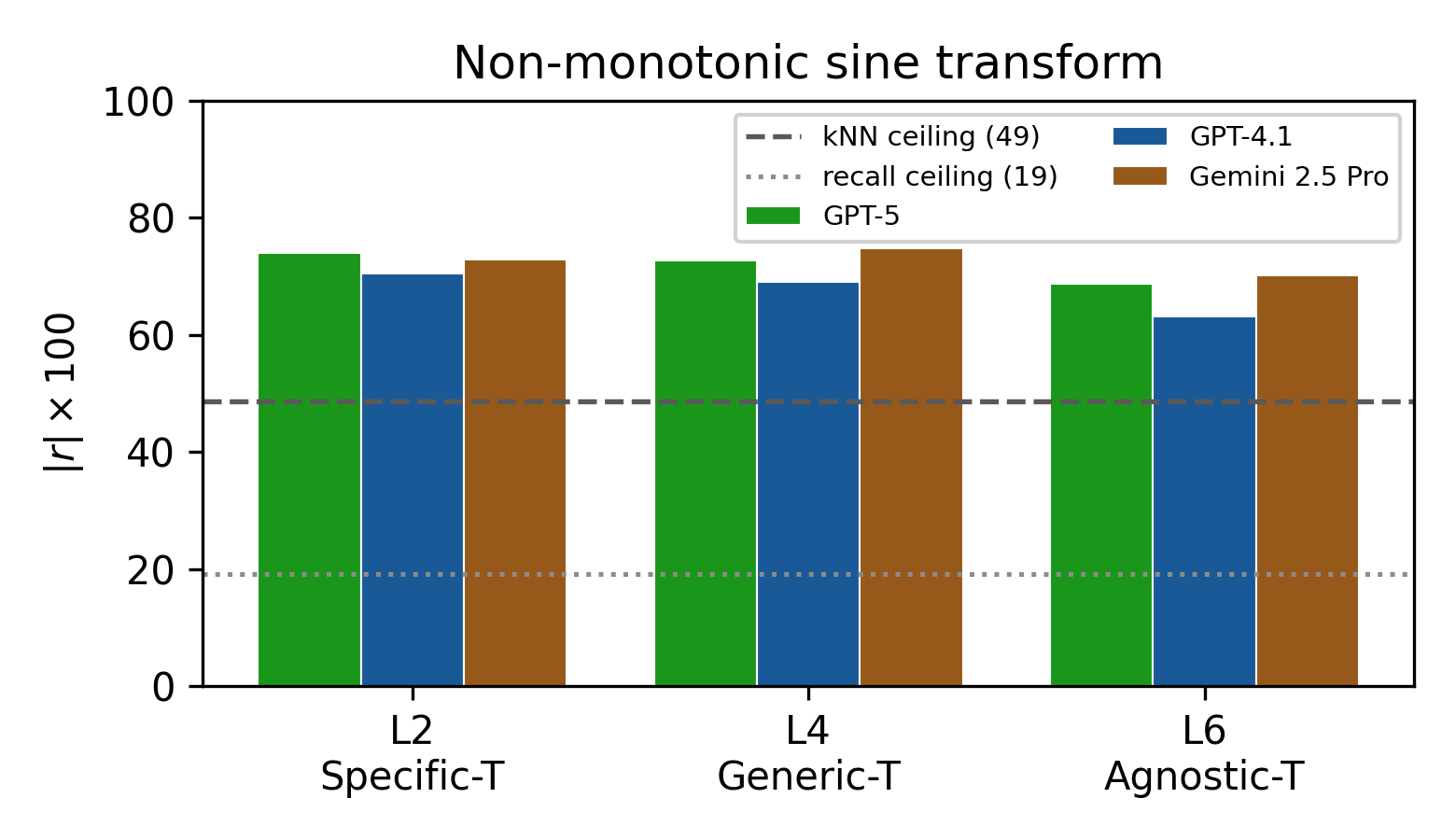}\caption{}\end{subfigure}
\caption{Non-monotonic transform control. Top: the continuous sine transform applied to the
labels (two full periods). Bottom: GPT-5, GPT-4.1, and Gemini~2.5~Pro $|r|\times100$ on Delaney
(1000-shot) at the three transformed blinding levels under (a) the affine, rank-preserving
transform and (b) the non-monotonic sine transform, with the kNN-Tanimoto structural ceiling
(dashed) and, in panel (b), the recall/prior-knowledge ceiling (dotted). Under the
non-recoverable sine relabeling every model stays well above both ceilings, evidencing genuine
in-context learning rather than scale recovery or value recall.}
\label{SI-fig:nonmono}
\end{figure}

An open question is whether our order-preserving (affine) value transform truly disrupts
recall, since a monotonic map can in principle be recovered from a few in-context anchors.
As a direct control we replaced it with a \emph{continuous non-monotonic} transform: the
min--max-normalised target $u=(y-y_{\min})/(y_{\max}-y_{\min})$ is mapped through a sine over
two full periods and rescaled to $[0,100]$,
\begin{equation*}
\tilde{y} = \tfrac{1}{2}\bigl(\sin(4\pi u)+1\bigr)\times 100 .
\end{equation*}
Unlike a binned permutation, this preserves local similarity everywhere (no
discontinuities), but it is globally non-monotonic and cannot be recovered by assuming
monotonicity from anchors. The price of continuity is that the map is many-to-one, so it
raises the intrinsic difficulty of the regression; we bound that with the kNN-Tanimoto
structural ceiling under the same transform. Over two periods, $\mathrm{Pearson}(y,\tilde
y)\approx-0.19$, so recall or prior knowledge of the true value is essentially useless for
$\tilde y$. We ran this control with GPT-5, GPT-4.1, and Gemini~2.5
Pro on Delaney at 1000-shot, over the same
150-molecule test split and two repetitions, at all three transformed blinding levels
(Specific-, Generic-, and Agnostic-Transformed; Table~\ref{SI-tab:nonmono}, Figure~\ref{SI-fig:nonmono}).

We highlight two observations, consistent across all three models and levels. First, every model reaches
$|r|=63$--$75$ under a non-recoverable transform whose target is uncorrelated with the true
solubility ($\approx$19 ceiling): this is well above the kNN structural ceiling of
$49$, so the LLMs evidently use a more accurate, problem-adapted notion of similarity than Tanimoto here.
Second, the sine transform makes the task substantially
harder for any structural model (the kNN ceiling drops $81\rightarrow49$, $-32$~pp), yet the LLM
$|r|$ falls only $\approx 17$--$23$~pp (paired-bootstrap CIs exclude zero in 8 of the 9 cells;
the sole exception is GPT-5 at the Generic level, whose affine baseline is itself anomalously low
at $68$). This complicates a direct comparison with our other results, but, relative to the kNN ceiling,
the picture remains qualitatively the same as under the linear transform.

\section{Reference Baselines}
\label{SI:baselines}

To gauge how much of each structure--property relationship is accessible without any memorization, we evaluate two classical baselines on the identical 150-molecule test split used for the LLMs.

\subsection{$k$-Nearest-Neighbour Tanimoto Baseline}

Using Morgan fingerprints (radius 2, 2048 bits), we predict the Tanimoto-similarity-weighted mean of the $k$ nearest training molecules' labels---the classical analogue of the algorithm the prompts ask the model to perform. Table~\ref{SI-tab:knn} reports the Pearson correlation ($\times100$) for a range of $k$ with 1000 training molecules. Because the correlation is invariant to the monotonic label transform, these values apply equally to transformed and untransformed targets.

\begin{table}[htbp]
\centering
\caption{kNN Tanimoto similarity-weighted-mean baseline, Pearson $r\times100$ vs.\ number of neighbours $k$ (1000 training molecules).}
\label{SI-tab:knn}
\small
\begin{tabular}{l|rrrrrrr}
\toprule
\textbf{Dataset} & $k{=}1$ & $k{=}3$ & $k{=}5$ & $k{=}10$ & $k{=}20$ & $k{=}40$ & $k{=}60$ \\
\midrule
Delaney & 76 & 80 & 81 & 81 & 79 & 74 & 74 \\
Lipophilicity & 40 & 47 & 49 & 43 & 42 & 45 & 43 \\
QM7 & 29 & 51 & 53 & 62 & 64 & 67 & 66 \\
\bottomrule
\end{tabular}
\end{table}

\subsection{Trivial-Feature Ridge Regression}

We fit ridge regression on two deliberately trivial feature sets derived from the SMILES string: per-element atom counts and a bag-of-characters of the raw string. Table~\ref{SI-tab:trivial} reports correlations at 60 and 1000 training examples, together with the single-feature correlation of the target with the heavy-atom count. Aqueous solubility (Delaney) is well captured by these trivial features, consistent with its near-linear structure--property relationship; QM7 atomization energy in this SMILES-only variant is strongly predicted by character counts (a size/composition proxy) but only weakly by heavy-atom count alone, unlike the canonical 3D QM7; Lipophilicity resists all trivial models.

\begin{table}[htbp]
\centering
\caption{Trivial-feature ridge regression, Pearson $r\times100$. ``elem'' = per-element atom counts, ``char'' = bag-of-characters of the SMILES; ``HAC'' = single-feature correlation with heavy-atom count.}
\label{SI-tab:trivial}
\small
\begin{tabular}{l|rr|rr|r}
\toprule
\textbf{Dataset} & elem-60 & elem-1000 & char-60 & char-1000 & HAC \\
\midrule
Delaney & 84 & 88 & 78 & 89 & $-61$ \\
Lipophilicity & 40 & 43 & 39 & 40 & 22 \\
QM7 & 68 & 68 & 88 & 90 & $-32$ \\
\bottomrule
\end{tabular}
\end{table}

\section{Bootstrap Confidence Intervals}
\label{SI:bootstrap}

Because we run two repeats and 150 test molecules per configuration, we quantify the test-set sampling uncertainty by bootstrapping over the 150 test molecules (resampling with replacement and recomputing the Pearson correlation; 5000 resamples). The median 95\% confidence-interval width is 13.3, 30.1, and 26.6 percentage points for Delaney, Lipophilicity, and QM7 respectively (overall median $\approx$26~pp). Consequently, small (few-point) differences between model variants are generally within noise, whereas the large effects emphasized in the main text are CI-separated. Table~\ref{SI-tab:bootstrap} illustrates this for Gemini~2.5~Pro across all six blinding levels (1000-shot; magnitudes $|r|$, with 95\% CIs).

\begin{table}[htbp]
\centering
\caption{Bootstrap 95\% confidence intervals for Gemini~2.5~Pro (1000-shot), $|r|\times100$ with $[\text{lower},\text{upper}]$, across blinding levels and datasets.}
\label{SI-tab:bootstrap}
\small
\begin{tabular}{l|ccc}
\toprule
\textbf{Blinding level} & \textbf{Delaney} & \textbf{Lipophilicity} & \textbf{QM7} \\
\midrule
Specific (L1) & 96 [94,97] & 72 [62,81] & 68 [56,80] \\
Spec-Transf (L2) & 96 [95,98] & 57 [41,71] & 71 [62,80] \\
Generic (L3) & 96 [95,97] & 65 [54,74] & 66 [53,78] \\
Gen-Transf (L4) & 96 [94,97] & 53 [39,65] & 75 [67,82] \\
Agnostic (L5) & 93 [90,96] & 42 [29,55] & 65 [54,75] \\
Agn-Transf (L6) & 93 [90,95] & 35 [22,48] & 71 [63,78] \\
\bottomrule
\end{tabular}
\end{table}

\subsection{Significance of the Directional Claims (Consistency Tests)}
\label{SI:significance}

The per-configuration intervals above show that most individual model-to-model
differences fall within test-set noise; we therefore make no fine-grained
model-ranking claims. The \emph{directional} effects the paper does assert can
nonetheless be tested jointly: an effect that is small in each model but points the
same way across the nine \emph{independent} models is collectively significant. Under
a null of no systematic effect each model's sign is a fair coin, so $k$ of $N$ models
agreeing is a one-sided binomial sign test against $p_0=0.5$; for contrasts evaluated
on the same 150 molecules (the six blinding levels) we additionally bootstrap the
paired difference $\Delta|r|$ (5000 resamples of the shared molecules), which is tighter
than comparing two marginal intervals. The $0$-shot runs use a separate, larger test
set, so $0$-shot contrasts use the unpaired sign test only. Table~\ref{SI-tab:significance}
collects the results; all correlations are computed with the paper's aggregation.

\begin{table}[htbp]
\centering
\caption{Significance of the main directional claims via consistency (sign) tests across
models. ``Cells'' counts the model$\times$dataset (or model) comparisons; $\overline{\Delta}$
is the mean change ($\times100$, percentage points); $p$ is the one-sided binomial sign test
on the direction. Effects that are individually within the $\sim$26~pp test-set CI become
jointly significant when they are consistent across the nine models.}
\label{SI-tab:significance}
\small
\begin{tabular}{l|l|c|r|c}
\toprule
\textbf{Claim} & \textbf{Contrast} & \textbf{Cells} & \textbf{$\overline{\Delta}$ (pp)} & \textbf{$p$} \\
\midrule
ICL helps (flagship models) & 1000-shot $>$ 0-shot & 8/9 & $+18.3$ & $0.020$ \\
\quad QM7 (all models) & 1000-shot $>$ 0-shot & 8/9 & $+29.2$ & $0.020$ \\
\quad Delaney (all models) & 1000-shot $>$ 0-shot & 7/9 & $+0.6$ & $0.090$ \\
\quad Lipophilicity (all models) & 1000-shot $>$ 0-shot & 2/9 & $-7.2$ & n.s. \\
\quad all models pooled & 1000-shot $>$ 0-shot & 17/27 & $+7.5$ & $0.124$ \\
Few-shot knowledge conflict & 60-shot $<$ 0-shot & 19/27 & $-3.9$ & $0.026$ \\
Context overcomes the dip & 1000-shot $>$ 60-shot & 21/27 & $+11.4$ & $0.003$ \\
Property name not decisive & Generic(L3) vs Specific(L1) & 14/27 & $-1.5$ & $0.50$ \\
Full blinding hurts & Specific(L1) $>$ Agnostic(L5) & 21/27 & $-8.6$ & $0.003$ \\
Lipophilicity collapse & L1 $>$ L6 & 8/9 & $+17.5$ & $0.020$ \\
\bottomrule
\end{tabular}
\end{table}

Three conclusions follow. First, the in-context-learning gain is real but \emph{not}
uniform. For the three flagship models the $0\!\to\!1000$-shot improvement is significant
($+18$~pp, $8/9$, $p=0.02$), and across all models it is large and significant on QM7
($+29$~pp, $p=0.02$). On Delaney the $0$-shot correlation is already near ceiling
($\sim$0.89), leaving little headroom; on Lipophilicity the $0$-shot prior-knowledge level
($\sim$0.4) already matches the structural in-context ceiling, so more context does not raise
it. Pooled over all nine models the $0\!\to\!1000$ change is therefore not significant
($17/27$, $p=0.12$)---which is why we restrict the improvement statement to the larger
models. Second, the few-shot knowledge conflict is supported: $60$-shot lies below $0$-shot in
$19/27$ cases ($p=0.026$) and is recovered by $1000$-shot ($21/27$, $p=0.003$); this U-shaped
behaviour is the clearest knowledge-conflict signature in the data.

Third, for the blinding levels we are deliberately conservative. Comparing the \emph{best}
of the six levels to Level~1 would be upward-biased (the maximum of six noisy levels exceeds
the first about five times in six by chance), so we test specific levels directly. Naming the
property generically (Level~3) is statistically indistinguishable from naming it specifically
(Level~1; $p=0.50$): the explicit property name is not what drives performance. Only \emph{full}
domain-agnostic blinding (Level~5) significantly lowers correlation ($-8.6$~pp, $p=0.003$), and
on Lipophilicity the drop to maximum blinding (L1$\to$L6) is large and consistent ($+17.5$~pp,
$8/9$ models, $p=0.02$; individually CI-separated for the three strongest models, e.g.\
Gemini~2.5~Pro $+37$~pp $[25,50]$ and GPT-5 $+40$~pp $[23,54]$). Delaney is robust: for the
flagship models the L1$\to$L6 change is only $3$--$6$~pp. A few individual cases are, however,
significant on their own and provide the clearest direct evidence that blinding can \emph{help}
by suppressing a misleading prior: most strikingly, Gemini~2.5~Flash on QM7 climbs from
$|r|\approx6\%$ at Level~1 to $\approx44\%$ at Level~6 ($+38$~pp, paired-bootstrap CI
$[+17,+54]$; the Level~4 transform alone already gives $+37$~pp, $[+13,+53]$), and Gemini~2.5
Flash-Lite on QM7 moves the same way ($+15$~pp) within noise. On Lipophilicity these two models
do \emph{not} show a significant blinding-induced gain (Flash in fact drops under full blinding).
Accordingly, the knowledge-conflict evidence rests on the significant $60$-shot dip, the
Lipophilicity asymmetry, and these specific significant blinding-improvement cases, rather than
on aggregate ``peak-above-Level-1'' differences, which are within test-set noise.

\section{Prompt Examples and Design Rationale}
\label{SI:prompts}

This section details the prompt templates used in our six-level blinding framework, using the Delaney (aqueous solubility) dataset as an example. Prompts for other datasets follow an identical structure, with only dataset-specific keywords adjusted in Levels 1 and 2. For each level, we provide the system prompt, analysis prompt (where applicable), and prediction prompt, accompanied by the rationale for the blinding strategy.

We adopt the prompting strategy from Busch et al., originally designed for 60 in-context examples. To accommodate 1000 examples while preserving their structured reasoning approach, we divide the context into two segments: 60 examples are retained in the analysis prompt to guide the reasoning process, while the remaining 940 examples are provided in a simplified sample list format (Name, SMILES, Property) inserted between the system and analysis prompts. This ensures efficient context utilization for large-scale in-context learning. We begin by presenting the 0-shot prompt.

\subsection{0-shot Prompt}

\subsubsection{System Prompt}

\begin{verbatim}
You are an expert chemist and know the delaney solubility
dataset very well. You are given a SMILES string of a molecule.
Your task is to predict the measured solubility of that molecule
in log solubility in mols per litre log(mol/L). Provide only the
numerical value as output, without any additional text.
\end{verbatim}

\textbf{Design Rationale:} This prompt provides maximum information:
\begin{itemize}
    \item \textbf{Role specification}: ``expert chemist'' primes chemistry-specific knowledge
    \item \textbf{Dataset}: ``delaney solubility dataset'' explicitly specifies the dataset
    \item \textbf{Explicit property}: ``log solubility of organic molecules in water'' leaves no ambiguity
    \item \textbf{Input format}: Clear identification of SMILES as chemical representation
    \item \textbf{Task framing}: Explicitly chemistry and solubility-focused
\end{itemize}

\subsubsection{Prediction Prompt}

\begin{verbatim}
What is the measured solubility in the delaney dataset in log(mol/L)
of the molecule with the following SMILES string: c1ccccc1? Provide
only the numerical value as output, without any additional text.
\end{verbatim}

\textbf{Design Rationale:} The prediction prompt is simple and direct, again specifying the exact property and the name of the dataset to give the LLM the chance to recall memorized values.

\subsection{Level 1: Specific (Baseline - Full Information)}

\subsubsection{System Prompt}

\begin{verbatim}
**Your Role**
You are a professional chemist with expert knowledge in
physical chemistry and solubility prediction. You are tasked
with predicting log solubility of organic molecules in water
based on their IUPAC names and SMILES strings.

**Problem Description**
You will be provided with:
1. A training dataset of molecules with their Names, SMILES,
   and known log solubility values (mol/L).
2. A test molecule (Name and SMILES) for which you must
   predict the log solubility.

You have to use your knowledge and abilities to analyze the
training data's molecules and the patterns and relationships
between molecular structures and solubility to make an
accurate prediction.
\end{verbatim}

\textbf{Design Rationale:} This prompt provides maximum information:
\begin{itemize}
    \item \textbf{Role specification}: ``professional chemist'' primes chemistry-specific knowledge
    \item \textbf{Explicit property}: ``log solubility of organic molecules in water'' leaves no ambiguity
    \item \textbf{Input format}: Clear identification of IUPAC names and SMILES as chemical representations
    \item \textbf{Task framing}: Explicitly chemistry and solubility-focused
\end{itemize}

\subsubsection{Analysis Prompt}

\begin{verbatim}
**Training Data:**
- Names: [benzene, toluene, phenol, ...]
- SMILES: [c1ccccc1, Cc1ccccc1, Oc1ccccc1, ...]
- Log Solubility (mol/L): [-1.52, -1.98, -0.82, ...]

**Analysis Task:**
1. Identify functional groups and structural features for
   each training sample.
2. Analyze how these features influence the solubility
   (e.g., hydrophilic vs hydrophobic groups).
3. Find pairs of similar molecules with different
   solubilities and explain the difference.

Think step by step and provide a systematic analysis of
the training data patterns.
\end{verbatim}

\textbf{Design Rationale:} The analysis phase uses chemistry-specific terminology (``functional groups,'' ``hydrophilic vs hydrophobic'') and explicitly mentions solubility. This encourages the LLM to retrieve and apply chemical knowledge.

\subsubsection{Prediction Prompt}

\begin{verbatim}
**Target Molecule:**
- Name: acetophenone
- SMILES: CC(=O)c1ccccc1

**Prediction Guide:**

**Similar Molecules Relations**
1. Similar Molecules: Find all similar molecules in the
   training data and analyze their relation to this compound
   with respect to the solubility process.
2. Similarity Analysis: Analyze similar molecules and rank
   them by similarity (with respect to mechanisms in the
   solubility process). Assign them a similarity value.

**Molecule Analysis**
3. Pattern Analysis: Analyze if found patterns apply
   to the molecule.
4. Functional Groups: Analyze its functional groups and
   their influence on the solubility.
5. Atomic Structure: Analyze its atomic structure and how
   this might influence the solubility.
6. Weighted Average: Calculate a weighted average of the
   solubilities of the similar molecules. Exclude molecules
   that have a small similarity value.

**Review and Prediction**
7. Prediction: Review your analysis shortly and write down
   the weighted average.
8. Result: As a result, write down one value and nothing
   after that. Syntax: "[Value]"
\end{verbatim}

\textbf{Design Rationale:} The 8-step structure provides explicit reasoning scaffolding. Steps 1-6 encourage analysis, step 7 synthesis, and step 8 ensures clean output formatting for automated parsing.

\subsection{Level 2: Specific-Transformed (Vague Hint, Transformed Values)}

\subsubsection{System Prompt}

\begin{verbatim}
**Your Role**
You are a professional chemist with expert knowledge in
physical chemistry and solubility prediction. You are tasked
with predicting a molecular property **related to solubility**
of organic molecules in water based on their IUPAC names and
SMILES strings.
\end{verbatim}

\textbf{Key Change from Level 1:} Property is now described as ``related to solubility'' (emphasis added) rather than explicitly ``log solubility.'' This subtle ambiguity tests whether the LLM can infer the exact property from examples.

\subsubsection{Analysis Prompt}

\begin{verbatim}
**Training Data:**
- Names: [benzene, toluene, phenol, ...]
- SMILES: [c1ccccc1, Cc1ccccc1, Oc1ccccc1, ...]
- molecular property related to solubility:
  [23.1, 30.7, 12.4, ...]   <- Transformed values

**Analysis Task:**
1. Identify functional groups and structural features for
   each training sample.
2. Analyze how these features influence the molecular
   property related to solubility (e.g., hydrophilic vs
   hydrophobic groups).
3. Find pairs of similar molecules with different molecular
   properties and explain the difference.
\end{verbatim}

\textbf{Key Change:} Values are mathematically transformed (multiplied by $-1$ and scaled to range $[0,100]$). This disrupts direct memorization while preserving rank order and functional relationships.

\subsection{Level 3: Generic (Generic Property, Clear Values)}

\subsubsection{System Prompt}

\begin{verbatim}
**Your Role**
You are a professional chemist with expert knowledge in
physical chemistry. You are tasked with predicting a
**molecular property** based on their IUPAC names and
SMILES strings.

[rest similar to Level 1, but "molecular property"
replaces "log solubility"]
\end{verbatim}

\textbf{Key Changes:}
\begin{itemize}
    \item \textbf{Generic property name}: ``molecular property'' (no mention of solubility)
    \item \textbf{Retained chemistry framing}: Still a ``professional chemist'' working with chemical inputs
    \item \textbf{Clear values}: Original (untransformed) solubility values used
\end{itemize}

This level tests whether chemical domain framing alone suffices when property identity is obscured.

\subsection{Level 4: Generic-Transformed (Generic Property, Transformed Values)}

\textbf{System Prompt:} Identical to Level 3.

\textbf{Analysis Prompt:} Same as Level 3 but with \textbf{transformed values}.

\textbf{Cumulative Blinding:}
\begin{itemize}
    \item (-) Property name removed
    \item (-) Values transformed
    \item (+) Chemistry context retained (``chemist,'' ``functional groups,'' SMILES/names)
\end{itemize}

This is the highest blinding level that retains chemistry-specific language.

\subsection{Level 5: Agnostic (Domain-Agnostic, Clear Values)}

\subsubsection{System Prompt}

\begin{verbatim}
**Your Role**
You are a professional **machine learning model** with expert
knowledge in **regression**. You are tasked with predicting a
**sample property** based on a **string based structure
representation** of the sample.

**Problem Description**
You will be provided with:
1. A training dataset of samples with their string based
   structure representation and known sample properties.
2. A test sample (string based structure representation) for
   which you must predict the sample property.

You have to use your knowledge and abilities to analyze the
training data's samples and the patterns and relationships
between sample properties to make an accurate prediction.
\end{verbatim}

\textbf{Key Changes (Complete Terminology Replacement):}
\begin{itemize}
    \item ``professional chemist'' $\rightarrow$ ``professional machine learning model''
    \item ``molecules'' $\rightarrow$ ``samples''
    \item ``SMILES'' $\rightarrow$ ``string based structure representation''
    \item ``molecular property'' $\rightarrow$ ``sample property''
    \item No chemistry-specific terms whatsoever
\end{itemize}

\subsubsection{Analysis Prompt}

\begin{verbatim}
**Training Data:**
- Sample structure strings: [araaaaar, AArAAAAAr, BArAAAAAr, ...]
- sample property: [-1.52, -1.98, -0.82, ...]

**Analysis Task:**
1. Identify **structural features** for each training sample.
2. Analyze how these features influence the sample property.
3. Find pairs of similar samples with different sample
   properties and explain the difference.
\end{verbatim}

\textbf{Design Rationale:} By removing all chemistry terminology, we test whether the LLM can extract patterns from pure string-based input representations without domain priming. ``Functional groups'' becomes ``structural features,'' maintaining task structure while eliminating chemical semantics.

\subsection{Level 6: Agnostic-Transformed (Maximum Blinding)}

\textbf{System Prompt:} Identical to Level 5.

\textbf{Analysis Prompt:} Same as Level 5 but with \textbf{transformed values} and \textbf{transformed SMILES strings}.

\textbf{Cumulative Blinding (Maximum):}
\begin{itemize}
    \item (-) No chemistry terminology
    \item (-) No property information
    \item (-) Transformed values
    \item (-) Transformed SMILES strings
    \item Generic regression framing only
\end{itemize}

This represents the most challenging scenario: predict a transformed property from string inputs with no domain knowledge. Success here would indicate robust pattern recognition capability.

\subsection{Design Principles Across All Levels}

Several design principles apply across all blinding levels:

\begin{enumerate}
    \item \textbf{Prompt structure preservation}: The 8-step prediction guide (identify similar examples, analyze patterns, weighted averaging) remains consistent across levels. Only the terminology changes. This ensures that performance differences reflect information availability rather than prompt quality.

    \item \textbf{Output formatting}: All levels use the same ``[Value]'' output format to enable automated parsing. The instruction to write the prediction last prevents parsing errors.

    \item \textbf{Chain-of-thought encouragement}: The multi-step structure and ``Think step by step'' instruction encourage explicit reasoning, making it easier to diagnose failure modes (did the model identify wrong similar molecules? wrong patterns? calculation errors?).

    \item \textbf{Minimal additional scaffolding}: We avoid providing excessive hints or examples beyond the training set. This tests the LLM's ability to extract patterns from the data alone.
\end{enumerate}

\end{document}